# Inpainting-Style Conditional Diffusion for Multivariable Time Series Forecasting

Kourosh Kiani[1,*] and S.M. Muyeen[2,*]
[1]Electrical and Computer Engineering, Semnan University, Semnan, Iran
[2] Electrical Engineering Department, Qatar University, Doha, Qatar
Corresponding Authors (kourosh.kiani@semnan.ac.ir, sm.muyeen@qu.edu.qa)

Abstract. Accurate solar power forecasting is essential for ensuring the reliable and efficient operation of modern power systems with high penetration of photovoltaic (PV) generation. However, the inherent intermittency, nonlinearity, and stochastic behavior of solar energy pose significant challenges for conventional forecasting approaches. In this paper, we propose a novel conditional diffusion-based framework for multivariable time-series solar power forecasting. The proposed method reformulates temporal PV data as structured two-dimensional representations (images) using a sliding-window patch construction, enabling the application of Denoising Diffusion Probabilistic Models (DDPM) within a unified spatiotemporal learning paradigm. A key contribution of this work is the formulation of solar forecasting as an inpainting problem, where future time steps are treated as missing regions to be reconstructed. This is achieved through a mask-based conditional diffusion mechanism, in which historical observations are preserved as conditioning context while the target (future) region is progressively corrupted and subsequently recovered via reverse diffusion. The model learns to generate coherent future sequences conditioned on observed data, effectively performing time-series inpainting. To fully utilize all available features and ensure compatibility with U-Net architectural constraints, a zero-padding strategy is introduced to construct fixed-size inputs. The model is trained using a supervised denoising objective to predict injected noise, enabling accurate iterative reconstruction during the reverse process. Extensive experiments conducted on benchmark PV dataset, including GEFCom2014, demonstrate that the proposed approach achieves high forecasting accuracy, particularly for short-term horizons. The results highlight the effectiveness of integrating diffusion-based generative modeling with an inpainting formulation for robust, flexible, and high-fidelity solar power forecasting.

## 1. INTRODUCTION

The increasing deployment of photovoltaic (PV) systems within modern power grids provides clear environmental benefits by reducing greenhouse gas emissions and decreasing dependence on fossil-fuel-based generation [1]. In addition to sustainability advantages, large-scale solar installations contribute to energy diversification and long-term economic savings [2]. However, the integration of high shares of solar power also introduces significant operational complexities, primarily because solar generation is inherently uncertain and highly dependent on rapidly changing meteorological conditions [3]. Fluctuations in solar irradiance, cloud movement, and atmospheric variability can result in sudden changes in power output, potentially causing mismatches between electricity supply and demand [4]. Consequently, accurate and reliable solar power forecasting has become a critical requirement for grid operators, energy market participants, and PV plant owners to support secure system operation, optimal scheduling, and efficient energy management [5]. Despite substantial research efforts, forecasting solar generation remains challenging due to its intermittent nature, strong temporal variability, and non-stationary statistical characteristics, which complicate the development of robust predictive models capable of capturing both short-term fluctuations and long-term patterns [6].

Over the past decade, numerous methodologies have been proposed to address the challenges associated with solar power forecasting. These approaches can generally be grouped into four main categories: statistical techniques [7], physics-based models [8], artificial intelligence–driven methods [9], and hybrid frameworks [10]. Statistical forecasting methods primarily exploit historical time-series measurements of PV generation along with related explanatory variables such as irradiance and temperature. Within this category, autoregressive (AR) [11], autoregressive moving average (ARMA) [12], The autoregressive integrated moving average (ARIMA) [13], and autoregressive moving average with exogenous inputs (ARMAX) [14] are the most-used models. Although these statistical methods are computationally efficient and relatively easy to interpret, their linear structure restricts their ability to capture the nonlinear and highly dynamic behavior that characterizes solar power output, particularly under rapidly changing weather conditions.

In contrast, physical forecasting approaches attempt to estimate PV production by explicitly modeling the relationships between atmospheric parameters, solar irradiance, and system characteristics [15]. Such methods rely on multiple data acquisition sources, including satellite-derived cloud imagery [16], measurements obtained from ground-based monitoring stations [17], and outputs generated by Numerical Weather Prediction (NWP) models [18]. While physics-based techniques can provide reliable predictions over certain forecast horizons, especially when high-quality meteorological inputs are available, they often involve complex preprocessing pipelines and substantial computational requirements [19]. Moreover, their predictive performance may degrade due to spatial and temporal uncertainties inherent in weather-related inputs, such as inaccuracies in estimating cloud movement, velocity, and optical thickness, all of which significantly affect solar irradiance reaching PV modules [20]. To alleviate these issues, clear-sky modeling strategies have been introduced to first estimate the theoretical irradiance under ideal atmospheric conditions and subsequently incorporate weather-induced deviations [21]. Several implementations have been reported in the literature, including the application of low-pass filtering for modeling clear-sky irradiance during daylight hours [22], the use of shortwave radiation time series combined with principal component analysis to characterize cloud variability [23], and approaches that employ statistical smoothing together with recursive least squares estimation to refine predictions [24]. Despite these advances, recent comparative studies suggest that data-driven artificial intelligence techniques are increasingly capable of achieving higher predictive accuracy than traditional physics-based approaches, particularly when large historical datasets are available for training [25].

Machine learning–based approaches have consequently attracted significant research interest for improving solar power forecasting performance [26]. These data-driven methods are generally categorized into two major groups: conventional (non-deep) machine learning algorithms and deep learning architectures. Conventional machine learning techniques include models such as multilayer perceptron (MLP) [27], support vector regression (SVR) [28], radial basis function (RBF) neural networks [29], gradient boosting machines (GBM) [30], and random forest (RF) models [31]. These algorithms typically rely on carefully designed input features derived from historical measurements, meteorological observations, and statistical transformations [26]. When appropriate feature engineering is performed, such methods can achieve satisfactory predictive accuracy and relatively low computational complexity. However, their effectiveness is strongly dependent on the quality and completeness of handcrafted features, which often requires extensive domain expertise. Moreover, these shallow models may struggle to fully capture the highly nonlinear, stochastic, and multi-scale temporal dynamics inherent in solar power generation, particularly under rapidly changing weather conditions or when long historical dependencies must be considered [27].

To overcome these limitations, deep learning–based forecasting methods have been increasingly explored due to their ability to automatically learn hierarchical feature representations directly from raw data [32]. Convolutional neural networks (CNN) [33] have been applied to extract spatial or temporal patterns from irradiance maps, satellite images, and sequential PV measurements. Recurrent neural network (RNN) variants, such as gated recurrent units (GRU) [34] and long short-term memory (LSTM) networks [35], are widely adopted for modeling temporal dependencies in sequential solar generation data and have demonstrated improved performance over traditional statistical methods in many forecasting scenarios. In addition, unsupervised architectures such as autoencoders (AE) [36] and deep belief networks (DBN) [37] have been used to learn compact latent representations that capture underlying structures within high-dimensional meteorological and generation datasets, which can subsequently enhance prediction accuracy when integrated with supervised forecasting models. More recently, Transformer-based architectures have emerged as a powerful alternative for time-series forecasting tasks, including solar power prediction [38]. By leveraging self-attention mechanisms, Transformers can effectively model long-range dependencies without the sequential processing constraints associated with recurrent networks. This capability allows them to capture complex temporal correlations across extended forecasting horizons [39]. Despite their advantages, standard Transformer models often require substantial computational resources and memory consumption, particularly when applied to long input sequences typical of high-resolution energy datasets [40]. To mitigate these limitations, several efficient Transformer variants have been proposed in the literature, including architectures designed to reduce attention complexity, memory usage, and training time while maintaining predictive performance. Examples include Informer [41], Reformer [42], and iTransformer [43], among others, which introduce mechanisms such as sparse attention, locality-sensitive hashing, and improved token representations to enhance scalability for long-sequence forecasting applications. Although these advanced deep learning models have demonstrated promising results, their performance remains sensitive to model configuration, training strategies, and hyperparameter selection.

Diffusion-based models have recently attracted increasing attention for time-series forecasting due to their strong generative capabilities and robustness in modeling complex stochastic processes. By progressively denoising latent representations through a predefined diffusion process, these models can effectively capture uncertainty, multi-modal distributions, and intricate temporal dynamics inherent in renewable energy generation [44,45]. For example, diffusion probabilistic models have been applied to probabilistic load and renewable energy forecasting, where they generate multiple plausible future trajectories to better quantify uncertainty. In particular, autoregressive diffusion frameworks such as TimeGrad have been proposed for multivariate probabilistic time-series forecasting, demonstrating improved performance over conventional deep learning baselines [46]. Similarly, score-based diffusion models have been extended to time-series modeling and imputation tasks, enabling robust learning of complex temporal distributions under uncertainty [47]. Furthermore, recent studies have explored diffusion-based approaches for energy and general time-series forecasting, showing their effectiveness in capturing stochastic patterns and improving probabilistic accuracy compared to traditional methods [48]. To enhance computational efficiency, latent diffusion architectures have also been introduced, which operate in compressed latent spaces while preserving generative fidelity, making them suitable for large-scale forecasting applications [49]. These developments highlight the growing potential of diffusion-based models for renewable energy forecasting. Nevertheless, challenges remain in terms of computational cost, training stability, and integration with domain-specific constraints, motivating further research into more efficient and adaptive diffusion-based forecasting frameworks.

To address the aforementioned limitations, increasing attention has been directed toward feature enrichment strategies that enhance the descriptive power of input datasets prior to model training. Feature enrichment aims to construct additional informative variables or representations that better capture latent patterns and dependencies embedded within PV generation data. These approaches commonly include the incorporation of exogenous meteorological variables such as temperature, humidity, wind speed, and cloud coverage [50], [51]. In addition, deep learning–based feature representation methods have recently been adopted in solar power forecasting. For example, convolutional neural networks (CNNs) have been utilized for spatial feature extraction from satellite and sky images [52], while recurrent neural networks, including LSTM and GRU models, have been applied to encode temporal dependencies and generate sequential feature representations [53]. Hybrid architectures such as CNN–LSTM models have also been explored to jointly capture spatial–temporal features [54]. Furthermore, unsupervised representation learning approaches, including autoencoders (AE) and variational autoencoders (VAE), have been employed to derive compact latent features from high-dimensional data [55]. More recently, Transformer-based embedding methods have been introduced to learn global contextual representations via self-attention mechanisms [38]. By enriching the input representation through these diverse strategies, forecasting systems can better characterize the variability and uncertainty inherent in solar generation, leading to improved robustness and generalization across different weather regimes, geographic locations, and forecasting horizons. Consequently, the design of effective feature enrichment pipelines has emerged as a critical research direction that complements advances in predictive modeling architectures. A systematic investigation of feature enhancement techniques is therefore essential for developing next-generation PV forecasting solutions capable of achieving higher accuracy and reliability under real-world operating conditions.

To tackle the challenges of the current works, in this paper, a novel framework is proposed, including the following contributions:

**Conditional diffusion-based forecasting via inpainting formulation:** Instead of using conventional direct regression or sequence-to-sequence forecasting models that predict future values in a deterministic manner, this method reformulates multivariate solar power forecasting as a time-series inpainting problem, where future values are treated as missing regions. A mask-guided conditional Denoising Diffusion Probabilistic Model (DDPM) is employed to progressively reconstruct these regions, enabling the model to generate coherent and context-consistent future sequences through a learned reverse diffusion process.

**Structured spatiotemporal representation with padding-aware design:** Instead of using raw one-dimensional time-series inputs or handcrafted feature engineering as in previous models, this method introduces a structured 2D patch-based representation combined with a zero-padding strategy to ensure compatibility with U-Net architectures. This design preserves all original features while enabling spatial–temporal learning, allowing the diffusion model to better capture inter-feature dependencies and temporal dynamics, ultimately improving forecasting performance across different prediction horizons.

The remainder of this paper is organized as follows. Section 2 presents the proposed diffusion-based forecasting framework. Section 3 describes the experimental setup, including implementation details, datasets, evaluation metrics, and comparative analyses, followed by a discussion of numerical results and performance evaluation. Finally, Section 4 concludes the paper and outlines potential directions for future research.

## 2. Proposed model

In this section, the details of the proposed model are presented. To facilitate this, we first provide an overview of both unconditional and conditional diffusion models, which serves as a foundation for a deeper understanding of the proposed approach.

## 2.1. Diffusion Model Framework

Unconditional Denoising Diffusion Probabilistic Model (DDPM) is a class of deep generative models consists of three conceptual phases as below:

**Phase 1 (Forward Unconditional Diffusion (Noising Process)):** The forward process gradually corrupts a clean image $X_0$ by adding Gaussian noise using time-dependent weights $\sqrt{\bar{\alpha}_t}$ and $\sqrt{(1-\bar{\alpha}_t)}$. At each time step, the image becomes a weighted combination of the original image and random noise. As the time step t increases, the cumulative scaling factor $\bar{\alpha}_t$ decreases, leading to a progressive reduction in the signal-to-noise ratio. Consequently, the structured information contained in the original image gradually vanishes, and the sample becomes increasingly dominated by Gaussian noise. At sufficiently large t, the distribution of $X_T$ approaches an isotropic Gaussian distribution, effectively erasing all semantic content of the original data. In this way, the forward diffusion process systematically transforms clean samples into nearly pure Gaussian noise, as illustrated in Figure 1.

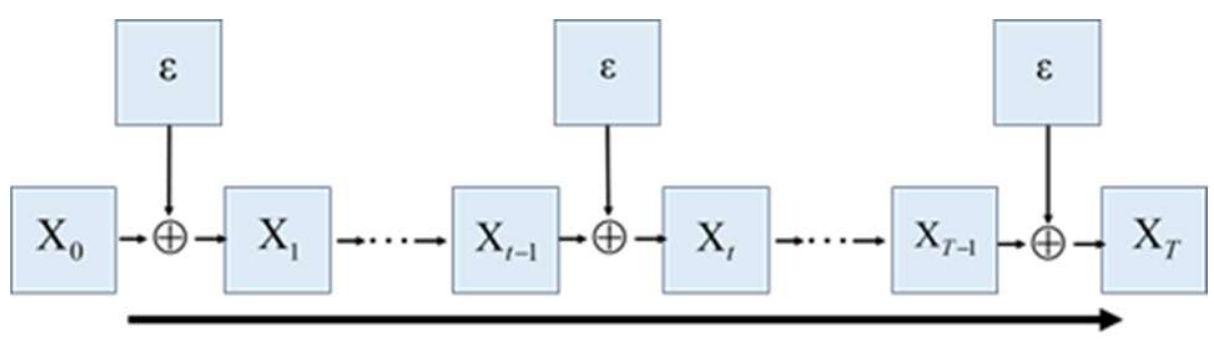

Figure 1: Forward Noising Process

With using the closed form forward diffusion equation as follows:

$$X_t = \sqrt{\bar{\alpha}_t}\, X_0 + \sqrt{(1-\bar{\alpha}_t)}\,\varepsilon \quad \varepsilon \sim \mathrm{N}(0,1), \tag{1}$$

we can sample noisy observations $X_t$ from clean data $X_0$ for arbitrary time steps and obtain the corresponding ground truth noise ε; these supervised pairs $(X_t, t, \varepsilon)$ constitute the training data for the denoiser neural network (See Phase 2). Furthermore, Figure 2 provides an intuitive visualization of the forward diffusion process using a portrait image as an example. This visual progression clearly demonstrates how the forward process systematically transforms structured data into an isotropic noise distribution. The following formulation provides an intuitive visualization of (1):

$$X_t = \sqrt{\bar{\alpha}_t} \quad + \sqrt{(1-\bar{\alpha}_t)} \tag{2}$$

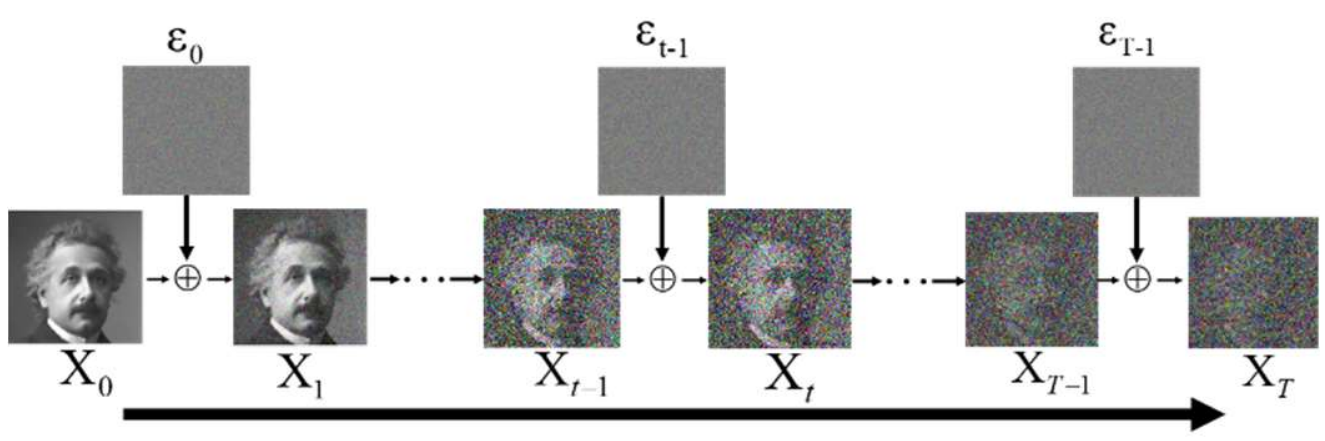

Figure 2: An intuitive visualization of the forward diffusion process

**Phase 2 (Training Unconditional Diffusion process):** It is not possible to do revers process by simply subtracting a known noise term, as shown in Figure 3, because the exact amount of noise added at each step is unknown during revers (inference) process. The reverse process begins from a randomly sampled Gaussian noise image $X_T \sim N(0,1)$. This sample is not obtained by progressively noising a specific clean image $X_0$. Consequently, there is no underlying forward trajectory and no record of the individual noise components that would have been added at each time step. Since no original image exists in this setting, the precise noise realization associated with any hypothetical forward process is unknown. The revers process (inference) cannot be computed by simply subtracting a known noise term and must instead be learned statistically. To solve this, the UNet model is trained to predict the injected noise $\varepsilon$ from the noisy image $X_t$ at time step t. During training, the same forward diffusion equation used to generate noisy samples is also employed to construct the supervision signal for the UNet. Specifically, a clean image $X_0$ is corrupted by adding Gaussian noise Ɛ at a randomly sampled time step t, producing the noisy input $X_0$. This noisy sample is then provided to the UNet together with the corresponding time step embedding. Crucially, the exact noise realization $\varepsilon$ that was injected through the forward process is retained and used as the training target. In this way, the model is optimized to predict the added noise from the corrupted input, enabling it to learn the time-dependent denoising function required for approximating the reverse diffusion dynamics. The most common loss in DDPM is the mean squared error (MSE) between the true noise and the predicted noise as follows:

$$\ell_{Uncondition} = \mathrm{E}_{X_0, \varepsilon, t}\left[\left\|\varepsilon - \varepsilon_\theta(X_t, t)\right\|^2\right] \tag{3}$$

where $\mathrm{E}_{X_0, \varepsilon, t}[\cdot]$ denote the expectation value function and $\theta$ represents the parameters of the predicted noise. Figure 5 shows an intuitive visualization of the reverse unconditional diffusion process. Additionally, Figure 4 provides a schematic illustration of the conditional diffusion mechanism employed during training.

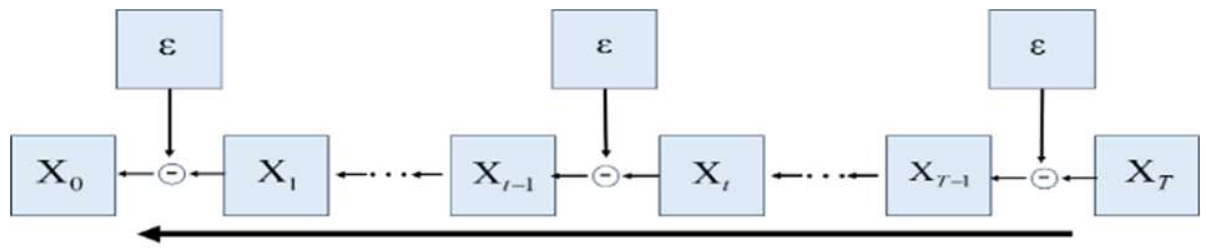

Figure 3: Reverse Denoising Process

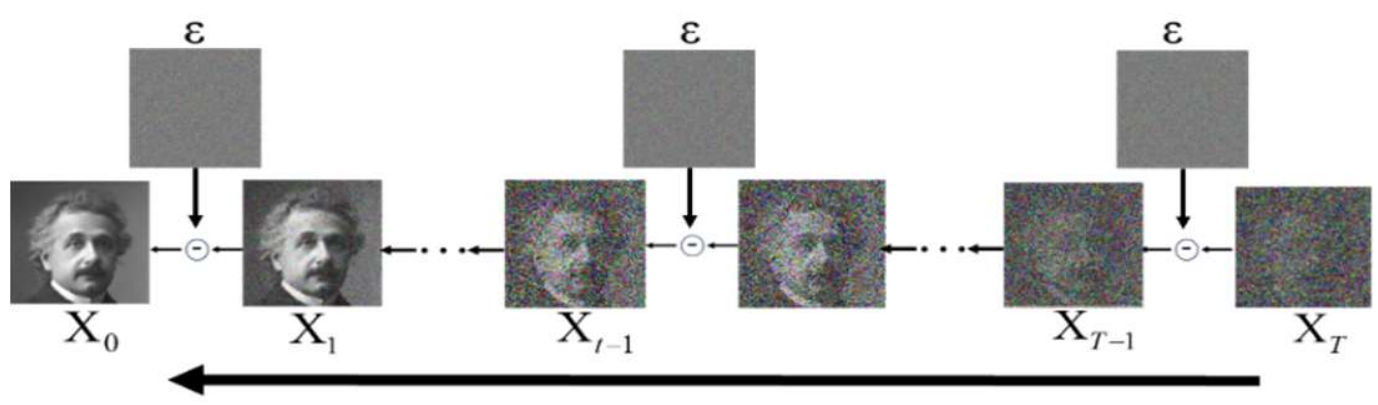

Figure 5: An intuitive visualization of the reverse diffusion process

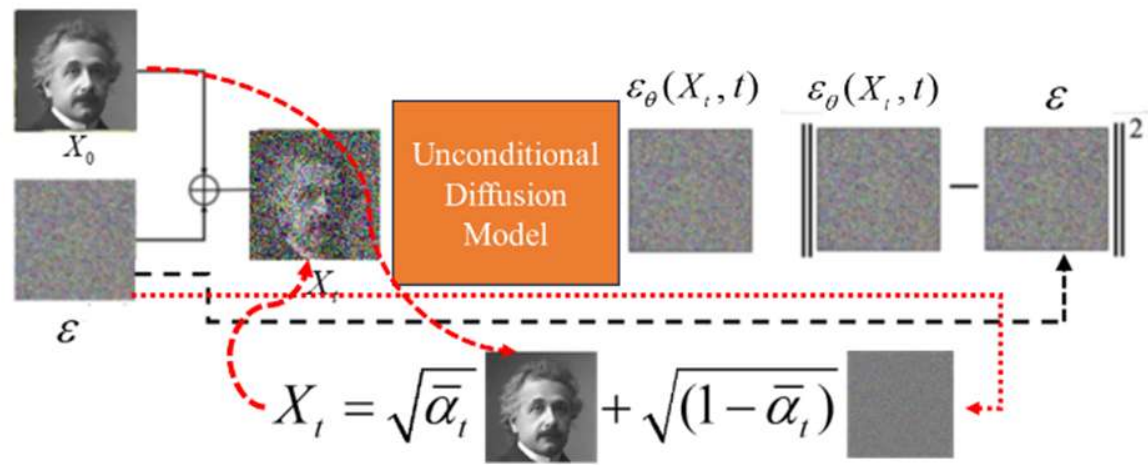

Figure 4: Training Unconditional Diffusion process

**Phase 3 (Reverse Unconditional Diffusion (Generation)):** After training, generation starts from pure noise $X_T \sim N(0,1)$ .The trained UNet estimates the noise at each time step, allowing iterative denoising from T down to 0. By progressively removing predicted noise, the model reconstructs a clean sample $X_0$, effectively reversing the forward diffusion process (See Figure 6).

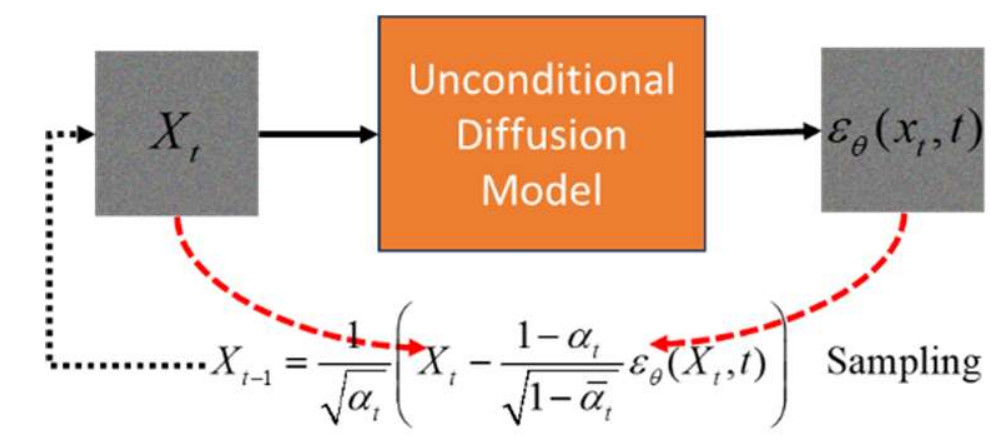

Figure 6: Iterative reverse Diffusion process from T down to 0.

In the example where only an image of Albert Einstein is used, the diffusion model learns to reconstruct and denoise that specific sample during training. However, if the training set is expanded to include portraits of Marie Curie, Stephen Hawking, Charles Darwin, and other well-known scientists (Figure 7), the model no longer learns a single instance but instead approximates the full data distribution p(x) over all of these individuals. As a result, the reverse diffusion process in an unconditional DDPM can converge to any of the learned models, meaning that sampling from noise may produce an image resembling any scientist present in the training set. This behavior reflects the model's attempt to represent the entire distribution p(x) rather than a single identity (See Figure ). Only when conditioning information is introduced, such as a class label or embedding corresponding to a specific scientist, does the reverse process become guided toward a particular target, ensuring that the generated image aligns with the desired identity rather than drifting among the various possibilities encoded in p(x).

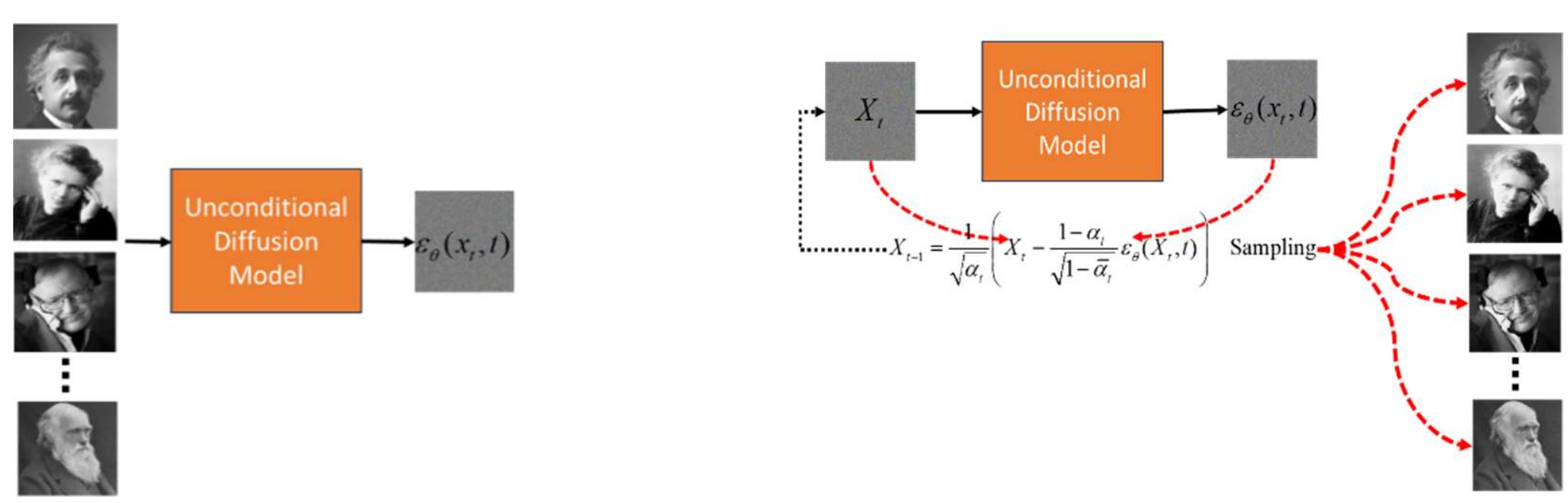

Figure 7: Training with many well-known scientists

Figure 8: Reverse outputs are randomly when DDPM is trained with many well-known scientists

**Phase 1 (Forward Conditional Diffusion (Noising Process)):** In the conditional Diffusion, the forward process remains structurally identical to the unconditional case. Specifically, the conditioning region corresponds to a fixed and non-learnable part of the image that remains unchanged throughout the diffusion process. The image is partitioned into two parts: a preserved conditional part, which is kept intact, and a target part to which Gaussian noise is progressively applied. During the forward diffusion process, only the target part is corrupted according to the standard noising equation, while the conditional part is directly provided to the model and is not modified or parameterized. As the time step increases, the target part gradually loses structural information and approaches a Gaussian distribution, whereas the preserved part maintains its original semantic content (See Figure 9). This design enforces that the model learns to reconstruct the noisy part in a manner that is consistent with the fixed contextual information, which is particularly suitable for tasks such as image inpainting or structured completion. The noising step at timestep t is formulated as follows:

$$X_t = (1-M)\odot X_0 + M\odot\sqrt{\bar{\alpha}_t}X_0 + M\odot\sqrt{1-\bar{\alpha}_t}\varepsilon \quad \varepsilon\sim \mathrm{N}(0,1) \tag{4}$$

where $X_0$ is the original clean input (e.g., a 16×16 patch of tabular time-series data), $M$ is a binary matrix (M = 1 in part to be noised/generated, M = 0 in conditioned/target part such as the top historical rows), $\odot$ denotes element-wise multiplication, $\varepsilon\sim \mathrm{N}(0,1)$ is standard Gaussian noise. $\sqrt{\bar{\alpha}_t}$ and $\sqrt{1-\bar{\alpha}_t}$ control the signal retention and noise injection strength at timestep t. Figure 10 shows a binary mask of size 16x16.

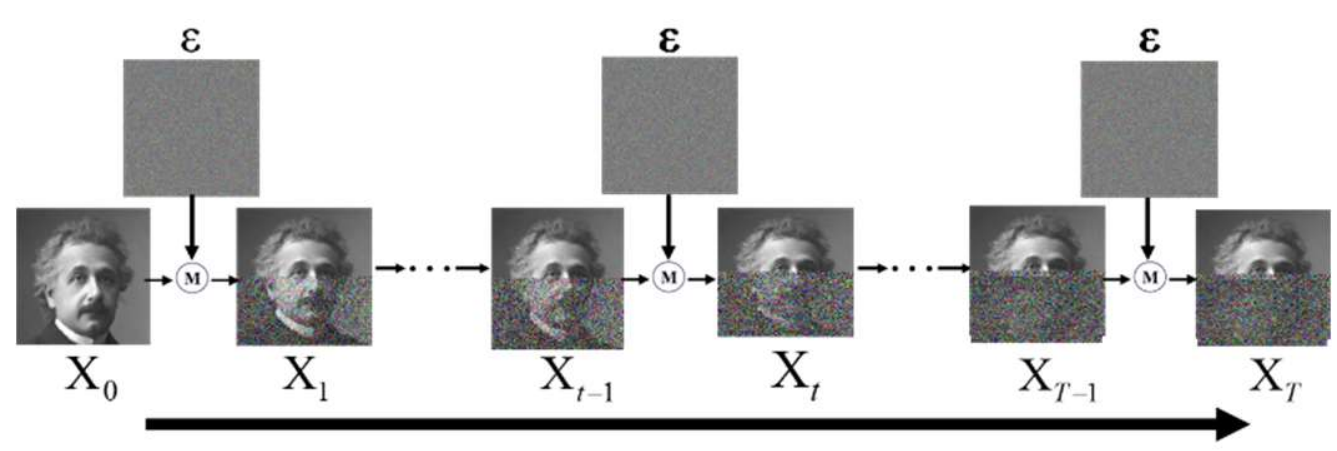

Figure 9: Forward conditional diffusion process

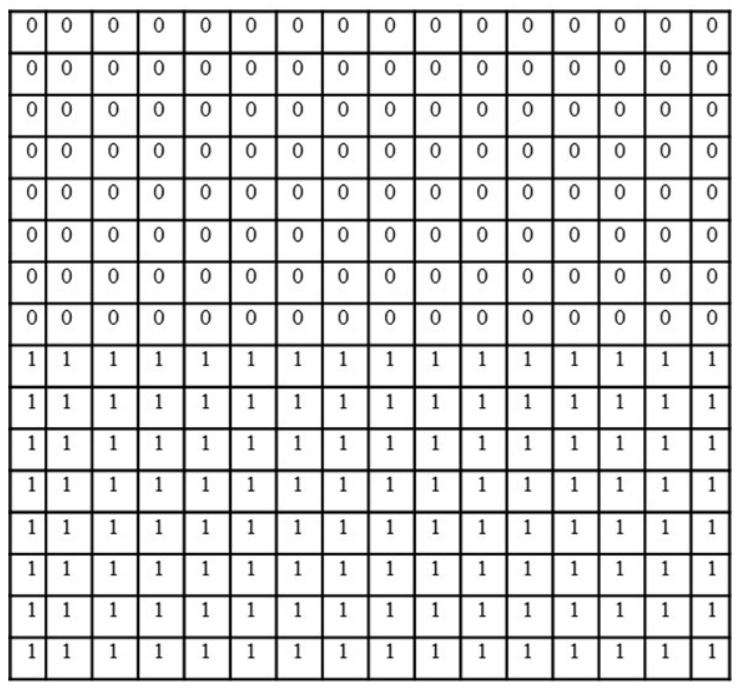

| 0 | 0 | 0 | 0 | 0 | 0 | 0 | 0 | 0 | 0 | 0 | 0 | 0 | 0 | 0 | 0 |
|---|---|---|---|---|---|---|---|---|---|---|---|---|---|---|---|
| 0 | 0 | 0 | 0 | 0 | 0 | 0 | 0 | 0 | 0 | 0 | 0 | 0 | 0 | 0 | 0 |
| 0 | 0 | 0 | 0 | 0 | 0 | 0 | 0 | 0 | 0 | 0 | 0 | 0 | 0 | 0 | 0 |
| 0 | 0 | 0 | 0 | 0 | 0 | 0 | 0 | 0 | 0 | 0 | 0 | 0 | 0 | 0 | 0 |
| 0 | 0 | 0 | 0 | 0 | 0 | 0 | 0 | 0 | 0 | 0 | 0 | 0 | 0 | 0 | 0 |
| 0 | 0 | 0 | 0 | 0 | 0 | 0 | 0 | 0 | 0 | 0 | 0 | 0 | 0 | 0 | 0 |
| 0 | 0 | 0 | 0 | 0 | 0 | 0 | 0 | 0 | 0 | 0 | 0 | 0 | 0 | 0 | 0 |
| 0 | 0 | 0 | 0 | 0 | 0 | 0 | 0 | 0 | 0 | 0 | 0 | 0 | 0 | 0 | 0 |
| 1 | 1 | 1 | 1 | 1 | 1 | 1 | 1 | 1 | 1 | 1 | 1 | 1 | 1 | 1 | 1 |
| 1 | 1 | 1 | 1 | 1 | 1 | 1 | 1 | 1 | 1 | 1 | 1 | 1 | 1 | 1 | 1 |
| 1 | 1 | 1 | 1 | 1 | 1 | 1 | 1 | 1 | 1 | 1 | 1 | 1 | 1 | 1 | 1 |
| 1 | 1 | 1 | 1 | 1 | 1 | 1 | 1 | 1 | 1 | 1 | 1 | 1 | 1 | 1 | 1 |
| 1 | 1 | 1 | 1 | 1 | 1 | 1 | 1 | 1 | 1 | 1 | 1 | 1 | 1 | 1 | 1 |
| 1 | 1 | 1 | 1 | 1 | 1 | 1 | 1 | 1 | 1 | 1 | 1 | 1 | 1 | 1 | 1 |
| 1 | 1 | 1 | 1 | 1 | 1 | 1 | 1 | 1 | 1 | 1 | 1 | 1 | 1 | 1 | 1 |
| 1 | 1 | 1 | 1 | 1 | 1 | 1 | 1 | 1 | 1 | 1 | 1 | 1 | 1 | 1 | 1 |

Figure 10: A binary mask(16x16)

Figure 11 shows a conceptual depiction of the forward conditional diffusion mechanism which is employed to generate training data during the UNet training conditional diffusion phase. A binary mask M (with values in {0,1}) is commonly employed during both the forward noising and reverse denoising processes to designate which part of the input remain fixed (conditioned) and which are subject to noise injection or generation. In the forward process, the masked formulation ensures that the conditioned part (typically the upper rows or known historical timesteps) retains its original clean values, while only the target part (e.g., future or missing rows) undergoes progressive corruption.

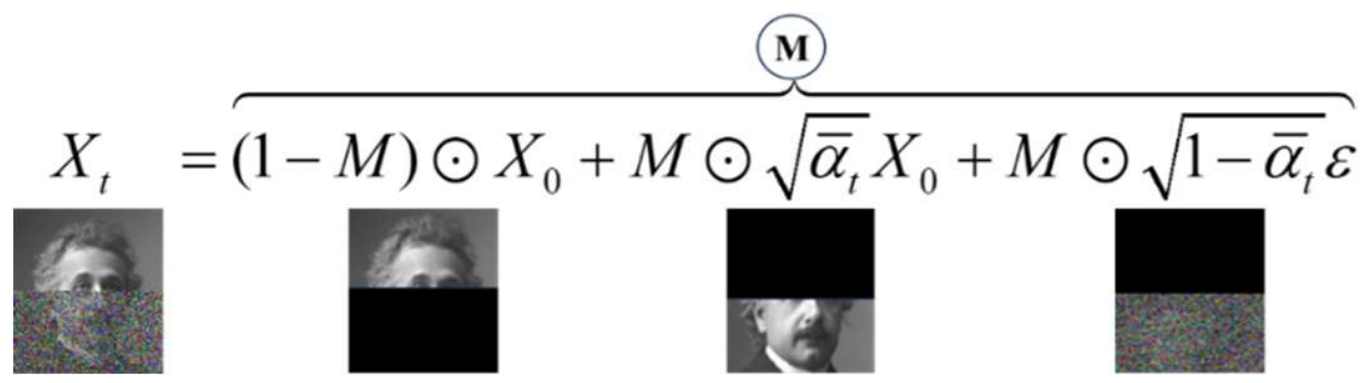

Figure 11: An intuitive visualization of the forward conditional diffusion process

**Phase 2 (Training Conditional Diffusion process):** During training of a conditional diffusion model, the forward noising process creates the noisy input $X_t$ and simultaneously provides the exact added noise $\varepsilon$ as the ground-truth label. This pair $(X_t, t, \varepsilon)$ is then used in a fully supervised regression manner to train the UNet to predict $\varepsilon_\theta(X_t, t, c)$ via mean squared error, making the entire learning procedure a supervised denoising task. Figure 12 shows a conceptual depiction of the training conditional diffusion phase.

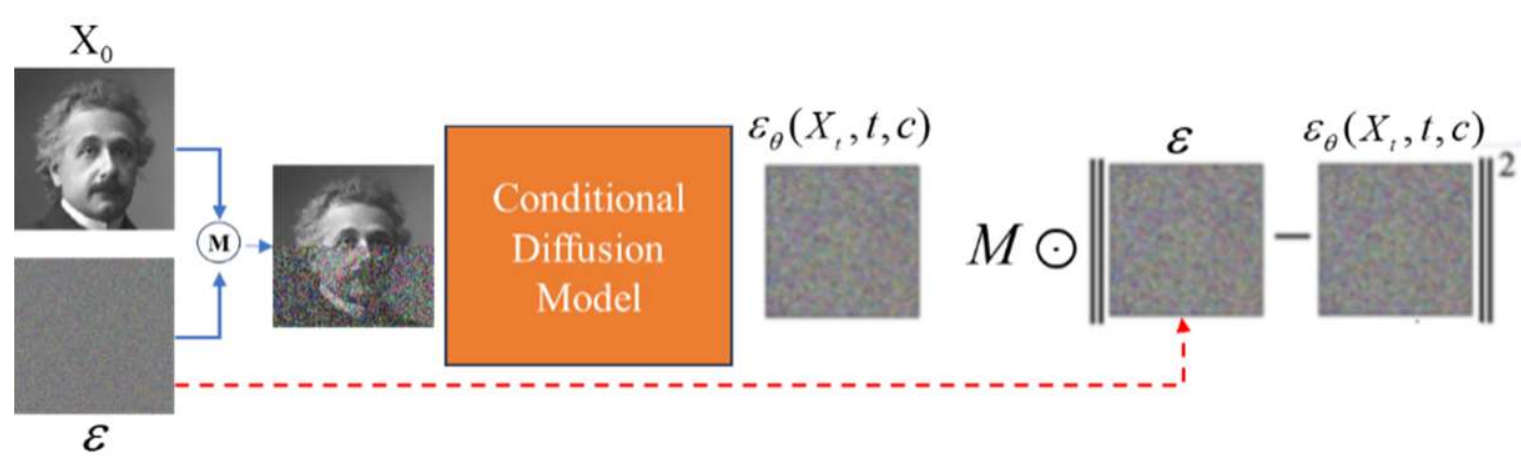

Figure 12: Training Conditional Diffusion process

The conditional noise prediction loss becomes:

$$\ell = \mathrm{E}_{X_0,\varepsilon,t,c}\left[\left\| M\odot(\varepsilon-\varepsilon_\theta(X_t, t, \overbrace{(1-M)\odot X_0}^{c}))\right\|^2\right] \tag{5}$$

where $X_t$ is the input of the conditional Diffusion model, $c$ represents the conditioning part, $\varepsilon_\theta(X_t, t, c)$ is the UNet prediction noise conditioned on $c$ , $M$ is binary mask matrix.

**Phase 3 (Reverse Conditional Diffusion (Generation)):** The reverse (generative) process in conditional diffusion begins with a partially noised state $X_T$, where the conditioned part (e.g., observed historical timesteps or top rows) remains clean and the target region is initialized with pure Gaussian noise $\varepsilon \sim \mathrm{N}(0,1)$. Guided by the conditioning part $c$, the trained UNet iteratively predicts the noise component $\varepsilon_\theta(X_t, t, c)$ at each timestep t, starting from t = T and proceeding backward to t = 0. At every step, the scheduler uses this predicted noise to compute a denoised estimate $X_{T-1}$, progressively refining the target part while preserving the fixed conditioned part. During sampling (i.e., the reverse denoising process), the same mask employed in the forward noising process is reused to maintain consistency between training and reverse process, ensuring the model focuses on learning to reverse-denoise only the target part. Through this controlled and iterative removal of predicted noise, the model reconstructs a coherent sample $X_0$ that remains faithful to both the learned data distribution and the provided conditioning, effectively reversing the masked forward diffusion process and enabling tasks such as time-series forecasting, and image inpainting. Figure 13 and Figure 14 show the reverse conditional diffusion. In contrast, an unconditional DDPM, which receives no conditioning part, generates a random plausible face. A conditional DDPM trained on a diverse dataset of scientists' portraits, the reverse process, initialized with the fixed conditioned part (e.g., the upper part of the image) remaining clean and the target region filled with pure Gaussian noise at t = T, iteratively denoises down to t = 0. This guided denoising reliably reconstructs the correct individual, accurately preserving identity, facial structure, and overall consistency with the conditioned context.

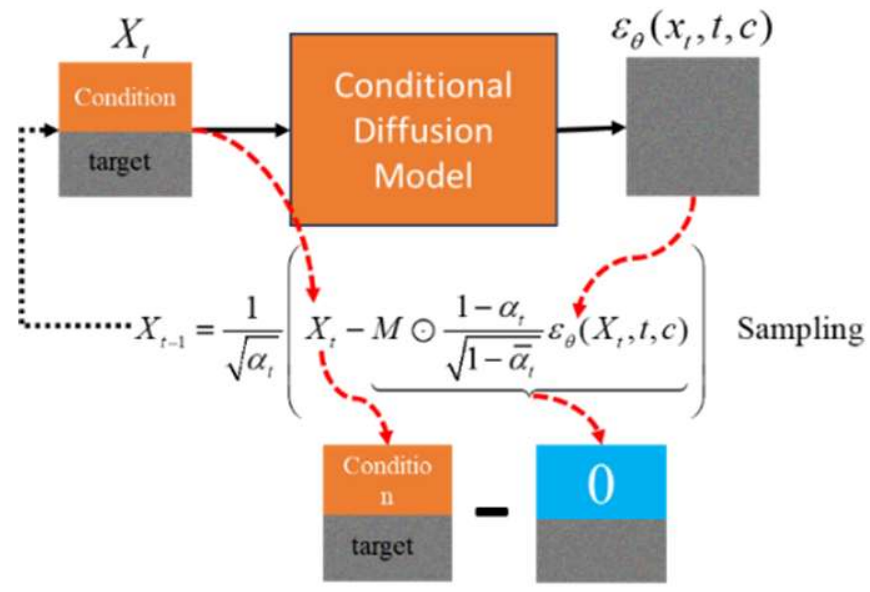

Figure 13: Reverse Conditional Diffusion

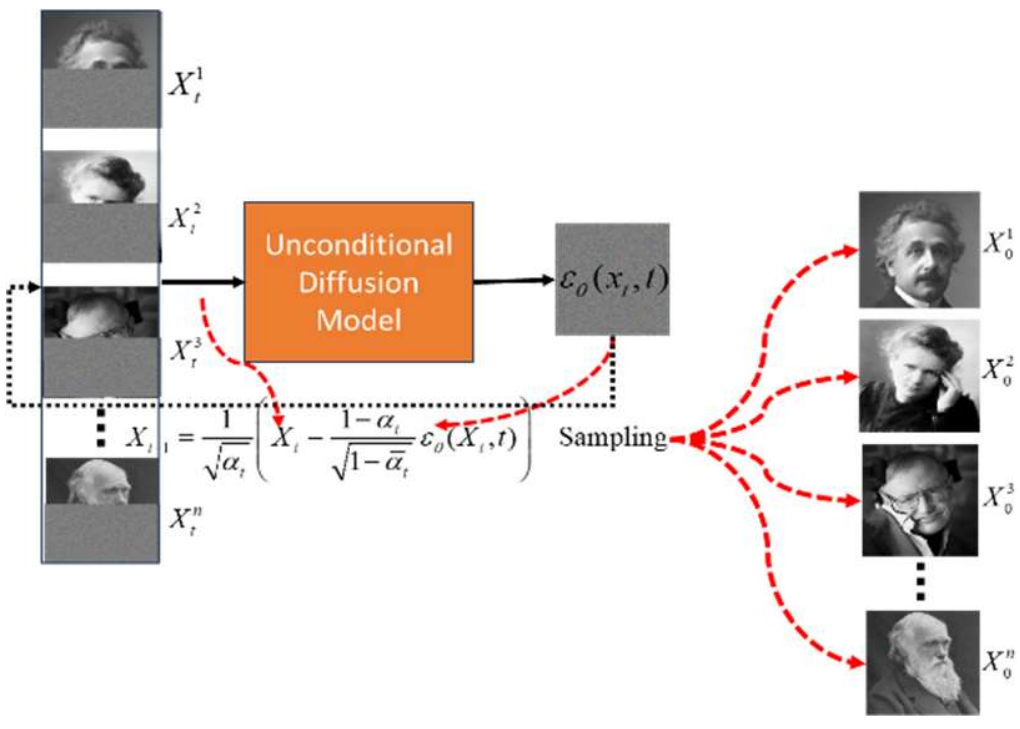

Figure 14: Reverse process of conditional DDPM reconstructs the correct individual

## 2.2. Data Representation and Patch Construction

The original dataset, denoted as GefCom2014, comprises a multivariate time series with 25 features recorded across 50375 sequential time steps. To enable conditional diffusion modeling on localized spatiotemporal patterns, only the first 16 features were selected, and the resulting sequence was transformed into fixed-size patches using a sliding-window procedure. A window of size 16×16 was applied, where each of the 16 columns corresponds to one of the selected features and each row represents a consecutive time step. The window was advanced along the temporal axis with a stride of one, producing a dense sequence of overlapping patches that collectively cover the full temporal evolution of the selected 16-feature subsequence. Each extracted window was treated as a single-channel image patch, allowing the diffusion model to operate in a spatially structured domain while preserving the inherent temporal ordering of the data. Figure 15 illustrates the construction of the first and second overlapping patches extracted from the dataset. A patch size of 16×16 was selected because the UNet architecture requires spatial dimensions that are consecutively divisible by 2 to enable efficient multi-scale feature extraction via repeated downsampling and upsampling. We later demonstrate how the patch size can be expanded to 32×32 to incorporate all 25 features while remaining fully compatible with the UNet's successive halving and doubling operations.

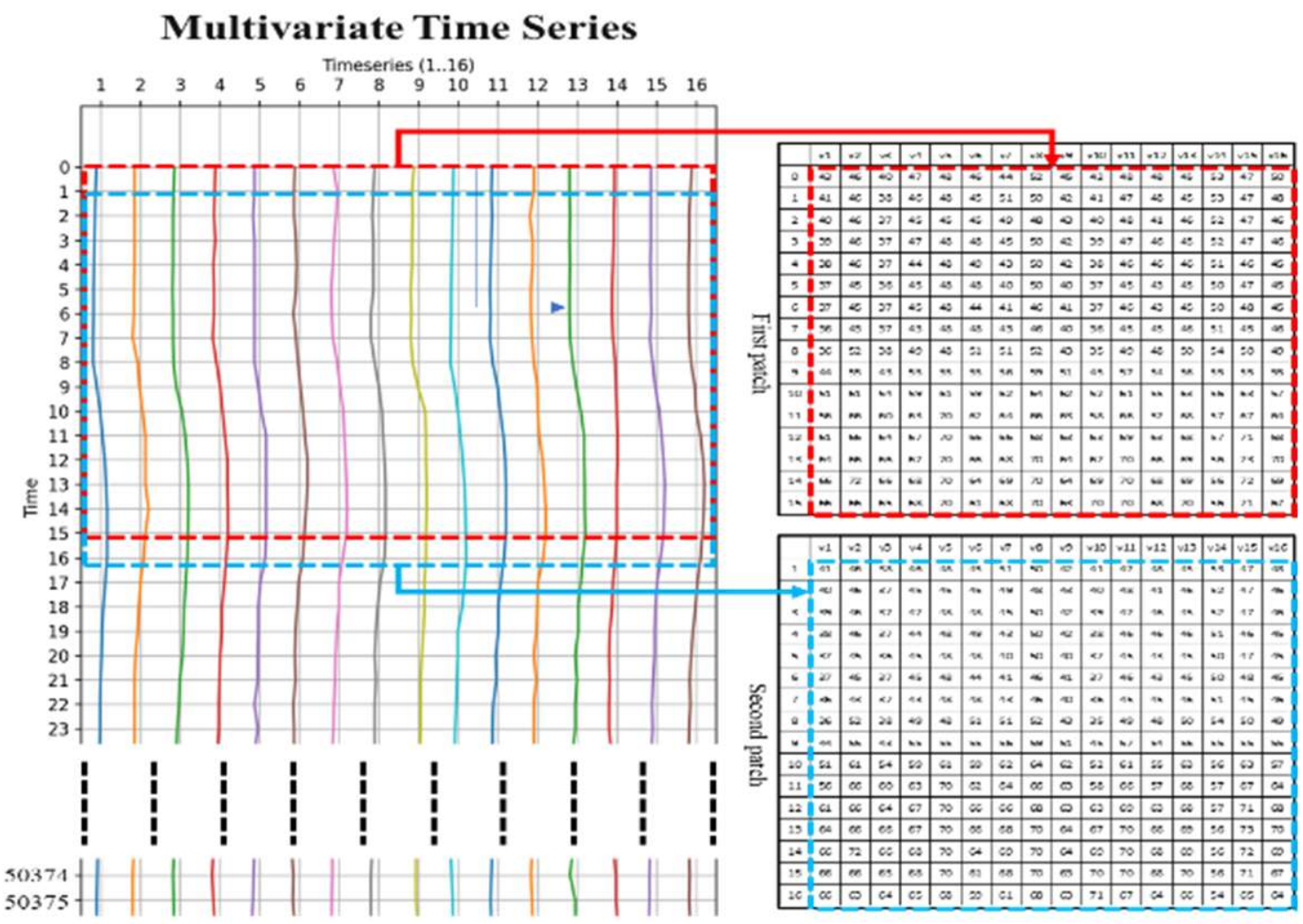

Figure 15: Construction of the first and second patches.

## 2.3. Condition–Target Partitioning for Conditional Diffusion

Figure 16 shows the partitioning of the 16×16 patch into conditioned (fixed) and target (generated) regions. The upper rows (white) serve as the conditioning context and remain unchanged, while the lower rows (gray) are noised during training and denoised during inference, enabling controlled prediction of future or missing timesteps conditioned on historical observations. It is worth mentioning that the patch size can be expanded to 32×32 to incorporate all 25 features while remaining fully compatible with the UNet's successive halving and doubling operations. To do this, the padding columns and rows are added when the number of features is not divisible by 2 (See Figure 17). Here, similar to the conditional diffusion model, there are three main phases as below:

**Phase 1 (Forward Conditional Diffusion process with padding):** To preserve all 25 features while satisfying the input-size constraints of the U-Net architecture, the multivariate time-series data are transformed into a two-dimensional image representation through zero-padding. The original matrix $X_t \in \mathbb{R}^{m \times 25}$ consists of 25 features (columns) and $m$ temporal samples (rows), where the number of rows is variable depending on the selected time window. Since the U-Net architecture performs multiple successive downsampling and upsampling operations, the input dimensions should be divisible by powers of two. Therefore, the feature dimension was fixed at 32 by appending 7 zero-valued columns: 25+7=32. The number of rows can vary and may contain up to 32 temporal samples without requiring padding. In our experiments, we selected 25 rows of temporal data, resulting in an initial 25 × 25 matrix. To construct a square 32 × 32 input image, 7 zero-valued rows were appended: 25+7=32. More generally, if fewer than 32 rows are selected, zero-padding is applied only to the remaining rows; if exactly 32 rows are selected, no row padding is needed. Thus, the padded input can be written as:

$$X_t^P = \text{padding}(X_t), \quad X_t^P \in \mathbb{R}^{32x\,32} \tag{17}$$

After constructing the 32 × 32 representation, the partitioning into conditional and target parts is performed only within the original data region and not within the padded area. In this partitioning, all 25 original features are retained and remain fixed across both the conditional and target parts, while the division is applied along the row dimension only.

The conditional rows correspond to the observed historical samples, and the target rows correspond to the future samples to be predicted. Thus, the model learns to generate future temporal patterns conditioned on past observations across the same fixed set of 25 features. The zero-padded rows and columns are excluded from prediction and serve only to satisfy the fixed input-size requirement of the U-Net architecture. This strategy preserves all original feature information, maintains the temporal ordering of the real data, and ensures compatibility with the hierarchical encoding and decoding operations of U-Net during both the forward conditional diffusion process and the reverse denoising stage.

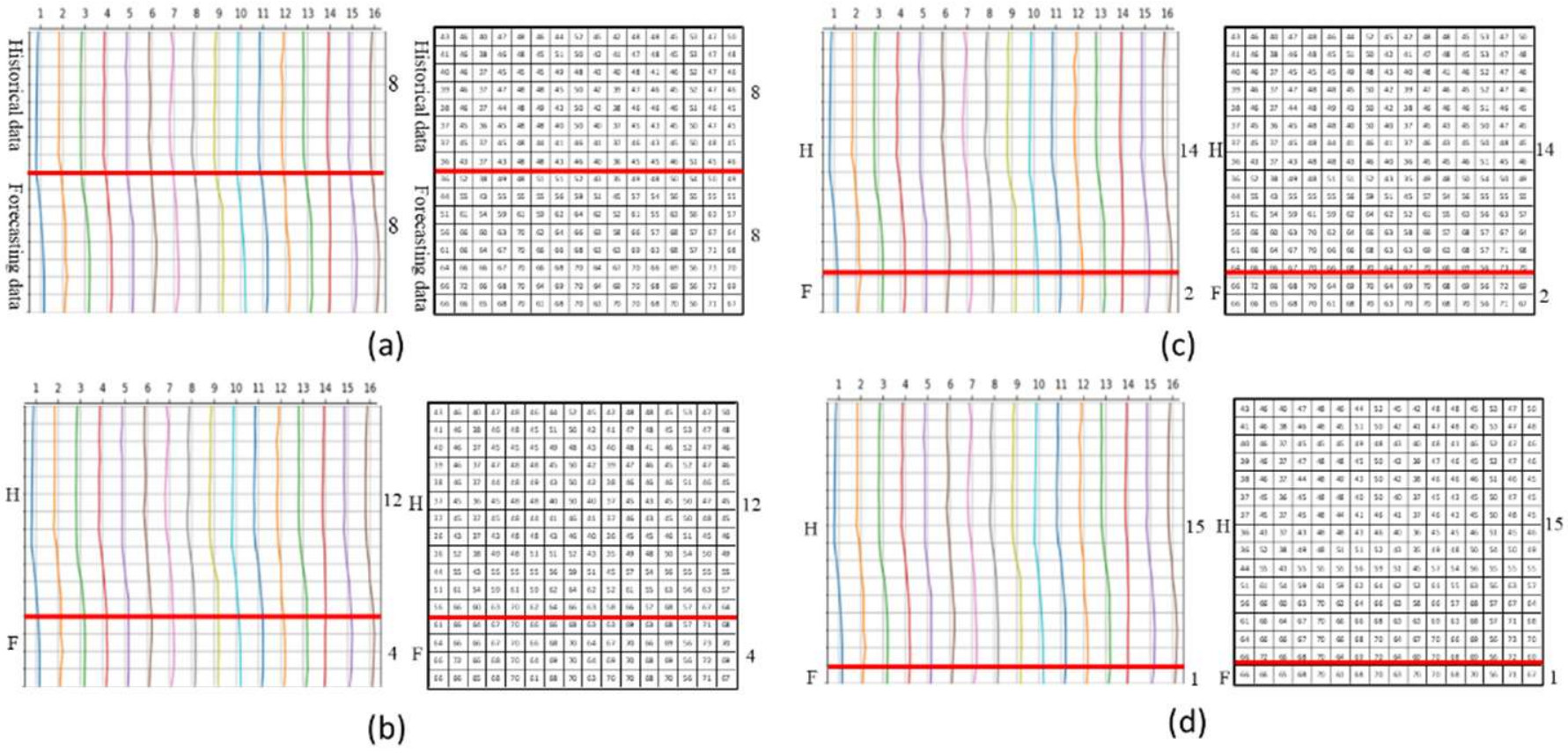

Figure 16: Partitioning of the 16×16 patch into conditioned and target: (a) 8 rows condition and 8 rows target. (b) 12 rows condition and 4 rows target. (c) 14 rows condition and 2 rows target. (d) 15 rows condition and 1 row target.

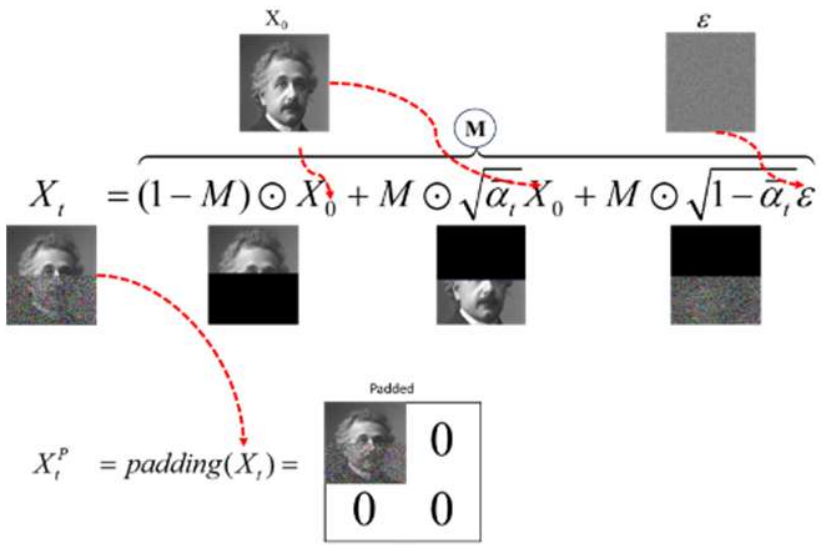

Figure 17: An overview of the padding mechanism.

**Phase 2 (Training Conditional Diffusion process with padding):** After constructing the padded representation, the original multivariate time-series image (matrix) $X_t$ is transformed into a fixed-size matrix $X_t^P$ through zero-padding before being provided as input to the U-Net model. The partitioning into conditional and target parts is performed only within the original data region of $X_t$, excluding the padded rows and columns. In this partitioning, all 25 original features remain fixed across both the conditional and target parts, while the split is applied only along the row dimension. The conditional rows represent the observed historical samples, and the target rows represent the future samples to be predicted. Thus, the model learns future temporal patterns conditioned on past observations over the same fixed feature space. The padded rows and columns do not participate in prediction and are included solely to satisfy the fixed-size input requirement of the U-Net architecture. During the training of the conditional diffusion model, the forward diffusion (noising) process is applied to the target region of $X_t$ by gradually adding Gaussian noise over a sequence of timesteps t. This process produces a noisy input $X_t$ and simultaneously records the exact added noise $\varepsilon$ as the ground-truth training label. The model parameters are optimized by minimizing the MSE between the true noise and the predicted noise. The output of the UNet model is unpadded:

$$\ell = \mathrm{E}_{X_0,\varepsilon,t}\left[\left\| M \odot (\varepsilon - \varepsilon_\theta(X_t, t, c)) \right\|^2\right] \tag{6}$$

where $\odot$ denotes element-wise multiplication. In this way, the network is explicitly optimized only to denoise and reconstruct the future target samples, while the conditional historical rows remain unchanged and serve purely as contextual guidance. This masked supervised denoising objective enables the model to learn the reverse diffusion process focused exclusively on future forecasting. Figure 18 conceptually illustrates this conditional diffusion training phase.

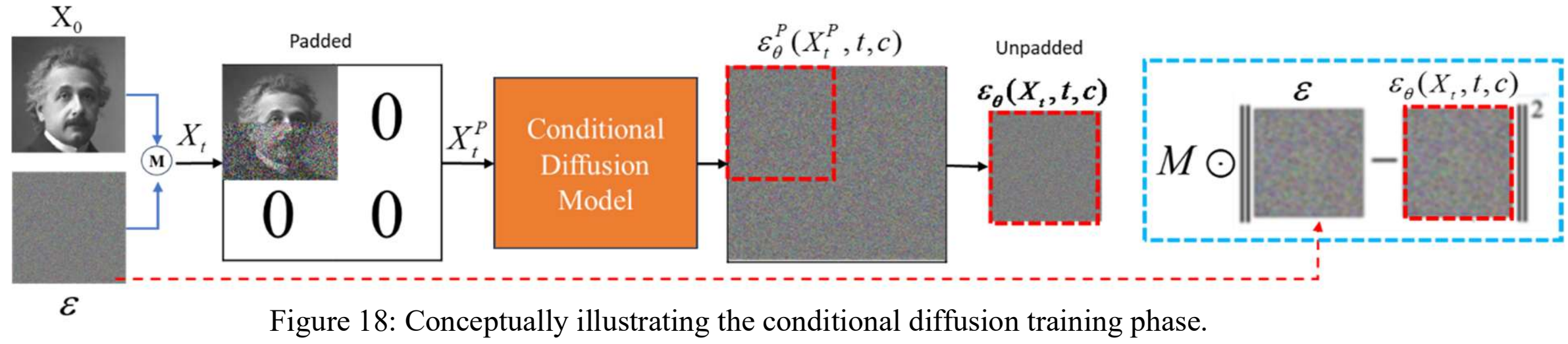

Figure 18: Conceptually illustrating the conditional diffusion training phase.

**Phase 3 (Reverse Conditional Diffusion with padding):** In the reverse conditional diffusion with padding phase, the forecasting process begins at the final diffusion timestep T, where the target region is initialized with pure Gaussian noise, while the conditional historical region remains fixed and unchanged. This noisy target region, together with the observed historical rows and the zero-padding, forms the 32 × 32 input representation fed into the U-Net model. Since the U-Net operates on the padded 32 × 32 representation, the predicted output is subsequently unpadded to recover the original valid data region (e.g., 25 × 25 in our experiments). By applying the mask, only the target region is progressively denoised, while the observed historical context remains fixed throughout the reverse process. Before, the $X_{t-1}$ being passed again to the U-Net for the next reverse timestep, it is padded again to reconstruct the fixed-size representation $X_{t-1}^{P}$. This iterative unpadding and repadding process ensures that only the meaningful data region is updated during sampling, while the network always receives an input of fixed size 32 × 32, compatible with the hierarchical encoding and decoding operations of U-Net. This procedure is repeated iteratively until t=0, progressively transforming the noisy target rows into realistic future samples conditioned on the fixed historical observations. In this way, the reverse conditional diffusion with padding process performs multi-step forecasting while keeping the correct time order of the data and ensuring that the input size always matches the required 32 × 32 format for the U-Net model. Figure 19 shows the reverse conditional diffusion process using padding for the U-Net input compatibility.

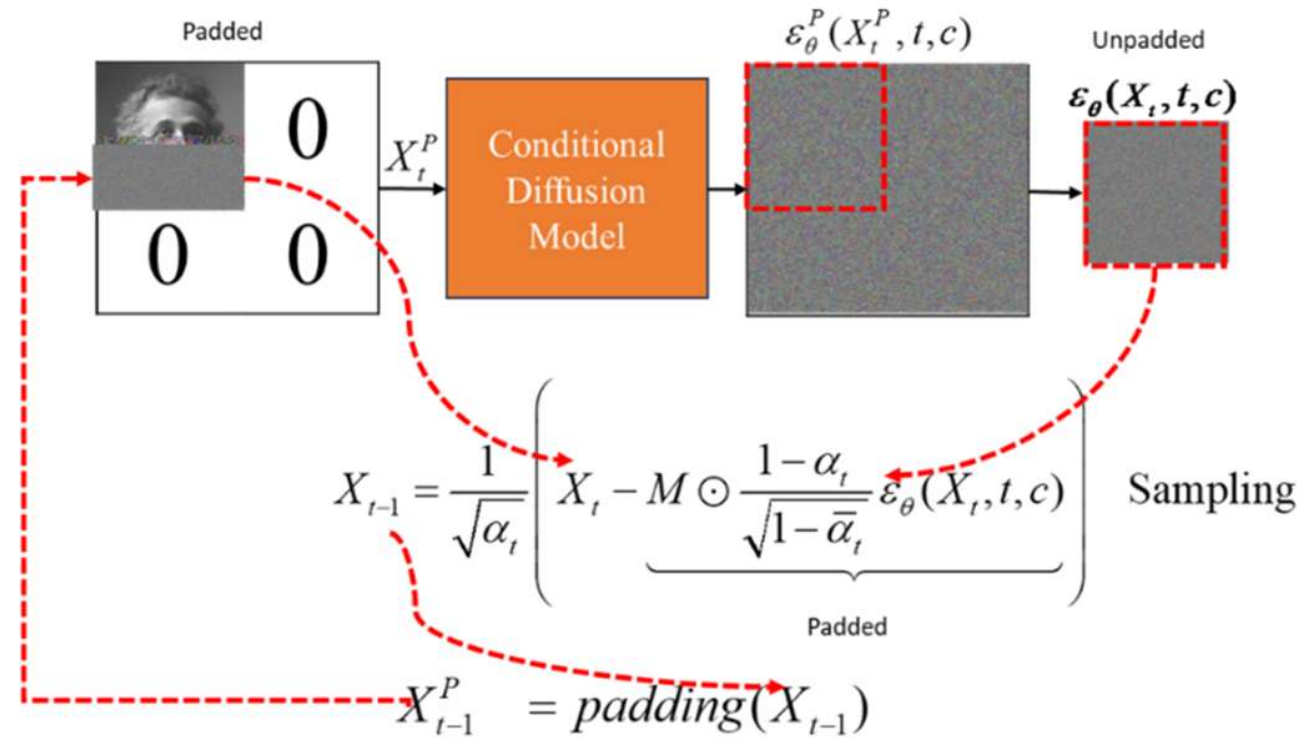

Figure 19: Reverse conditional diffusion process using padding for x × x U-Net input compatibility.

## 3. EXPERIMENTAL RESULTS

In this section, details of the implementation, datasets, comparative models, and numerical results are explained.

### 3.1. Implementation details

In this section, we provide a comprehensive presentation of the numerical evaluation conducted to assess the effectiveness of the proposed solar power forecasting framework. All experimental analyses were carried out using a high-performance computing environment configured with two Intel® Xeon® E5-2699 processors and 50 GB of system memory, operating under the Microsoft Windows 10 platform. To accelerate the training and inference procedures of the deep learning components, an NVIDIA Tesla K80 graphical processing unit (GPU) was employed. This hardware configuration ensured efficient handling of large-scale time-series datasets and computationally intensive optimization routines associated with the proposed methodology.

The implementation of the forecasting pipeline was developed using the Python programming environment, which provided the flexibility required for integrating multiple algorithmic modules within a unified workflow. All model training, validation, and testing procedures were executed under consistent experimental settings to ensure reproducibility and fair performance assessment. Hyperparameter tuning, optimization, and model convergence were monitored throughout the training process to guarantee stable learning behavior and reliable evaluation of forecasting accuracy. This computational setup allowed the proposed framework to be rigorously tested across multiple experimental scenarios while maintaining scalability and efficient resource utilization.

### 3.2. Datasets

To comprehensively evaluate the performance and generalization capability of the proposed solar power forecasting framework, a real-world solar power dataset is employed in the numerical experiments. This dataset is publicly available and has been widely used in the literature, enabling reproducible evaluation and fair comparison with existing approaches. This dataset is derived from the Global Energy Forecasting Competition 2014 (GEFCom2014) [56], which represents one of the most established benchmark datasets for solar power forecasting research. This dataset provides hourly PV power generation measurements along with corresponding forecasts of twelve meteorological variables for three geographically distinct solar power plants located in Australia, referred to as Zone 1, Zone 2, and Zone 3. The data span a continuous period from April 2012 to May 2014 and are publicly accessible through the official GEFCom repository. In this study, we focus on the solar power plant located in Zone 1, which has a nominal capacity of 1.56 kW. This subset is selected due to its frequent use in prior studies, allowing direct performance comparison and facilitating benchmarking against state-of-the-art forecasting models. The dataset's high temporal resolution and rich meteorological information make it particularly suitable for assessing short-term and day-ahead forecasting accuracy.

For hourly solar power prediction, a preprocessing procedure is applied to account for the intrinsic diurnal pattern of solar irradiance. Specifically, each test day is divided into day-time hours, during which solar radiation is present, and night-time hours, during which solar generation is physically absent. The partitioning of day-time and night-time intervals for a given test day is determined based on the solar radiation profile of the preceding day. This assumption is justified by the fact that sunrise and sunset times typically evolve gradually over longer temporal scales, such as seasonal transitions, rather than exhibiting abrupt changes between consecutive days. Accordingly, solar power values corresponding to night-time hours are set to zero, while the proposed forecasting framework is applied exclusively to predict PV output during the identified day-time hours. This strategy ensures physically consistent forecasts and prevents the model from learning spurious patterns during periods of zero generation. This preprocessing approach, combined with the diversity of datasets used in the evaluation, enables a rigorous assessment of the proposed method's ability to capture both short-term variability and longer-term temporal structures in solar power generation under different climatic and operational conditions.

### 3.3. Evaluation metrics

To assess solar power prediction performance of different methods, three widely-used error criteria, including Mean Absolute Percentage Error (MAPE), Mean Squared Error (MSE), and Mean Absolute Error (MAE) are employed in this section [59]:

$$MAPE(\%) = \frac{1}{N}\sum_{i=1}^{N}\frac{\left|Y_i^{true} - Y_i^{forecast}\right|}{\bar{Y}^{true}} \times 100 \tag{7}$$

$$\bar{Y}^{true} = \frac{1}{N}\sum_{i=1}^{N} Y_i^{true} \tag{8}$$

$$MSE = \frac{1}{N}\sum_{i=1}^{N}(Y_i^{true} - Y_i^{forecast})^2 \tag{9}$$

$$MAE = \frac{1}{N}\sum_{i=1}^{N}\left|Y_i^{true} - Y_i^{forecast}\right| \tag{10}$$

$$Pearson = \frac{\sum_{i=1}^{N}(Y_i^{true} - \bar{Y}^{true})(Y_i^{forecast} - \bar{Y}^{forecast})}{\sqrt{\sum_{i=1}^{N}(Y_i^{true} - \bar{Y}^{true})^2}\sqrt{\sum_{i=1}^{N}(Y_i^{forecast} - \bar{Y}^{forecast})^2}} \tag{11}$$

where $N$ indicates the number of day-time hours of the test day; $Y_i^{true}$ and $Y_i^{forecast}$ represent the actual solar power value and forecasted solar power value of hour $i$ in the test day, respectively; and $\bar{Y}^{true}$ is the mean of the hourly actual solar power values of the test day. $\bar{Y}^{true}$ is used in the MAPE's denominator in (15) to prevent the numerical problem caused by zero/close-to-zero actual hourly values. The Pearson correlation coefficient evaluates the linear correlation between the actual and predicted solar power sequences. A Pearson value closer to 1 indicates stronger agreement and higher forecasting accuracy.

### 3.4. Numerical results

In this section, results of the proposed framework are presented and discussed in two main categories: ablation analysis and state-of-the-art results.

**Ablation analysis**: Table 1 summarizes the performance of the proposed U-Net-based conditional diffusion model under different image-construction strategies for transforming multivariate time-series data into a two-dimensional representation suitable for image-based forecasting. The original dataset consists of 25 temporal features. Since the U-Net architecture requires square inputs with dimensions divisible by powers of two, multiple reshaping and padding strategies were systematically investigated to evaluate their impact on forecasting performance. In Experiments 1 to 4, only 16 out of the 25 available features were selected and arranged into a 16 × 16 representation without any padding. Within this setting, the number of conditional rows (i.e., historical observations used as input) was progressively reduced from 15 to 8, while the number of target rows (i.e., future time steps to be predicted) increased from 1 to 8. This configuration allows for isolating the effect of prediction horizon while maintaining a fixed spatial input size. The results show a clear and consistent trend: forecasting accuracy degrades as the prediction horizon increases. Specifically, the Pearson correlation decreases slightly (from 0.992 to 0.989), while both MAP and RMSE increase noticeably, indicating that longer-term predictions introduce greater uncertainty and are more difficult for the model to reconstruct accurately. In Experiments 5 to 8, all 25 features were retained and initially organized into a 25 × 25 representation. To satisfy the architectural constraints of the U-Net, symmetric zero-padding was applied by adding 7 rows and 7 columns, resulting in a 32 × 32 input. Compared to the first group of experiments, this configuration provides a richer and more complete feature space, allowing the model to exploit full inter-feature dependencies. Similar to the previous setup, the prediction horizon was gradually increased from 1 to 8 target rows. The results demonstrate a consistent improvement over Experiments 1 to 4 across all metrics. In particular, the higher Pearson correlation and lower MAP and RMSE values indicate that preserving all features significantly enhances the model's ability to capture both temporal dynamics and cross-feature relationships. This highlights the importance of comprehensive feature utilization in diffusion-based forecasting frameworks. In Experiments 9 to 12, an alternative padding strategy was explored in which only the feature dimension (columns) was padded to 32, while the temporal dimension (rows) was extended to 32 without symmetric padding. This design preserves higher temporal resolution and provides the model with a longer historical context. The results in this group are competitive with those of Experiments 5 to 8, particularly for short-term forecasting (1–2 target rows), where performance remains very strong. However, for longer horizons (4–8 target rows), a slight degradation is observed compared to the symmetric padding approach. This suggests that while increased temporal resolution is beneficial, maintaining a balanced and structured spatial representation may be more effective for stable long-horizon predictions within the U-Net-based diffusion framework.

Across all experimental configurations, short-term forecasting (1–2 target rows) consistently achieves the best performance, with the highest Pearson correlation and the lowest MAP and RMSE values. As the prediction horizon increases to 4 and 8 rows, performance gradually deteriorates. This behavior is expected, as the diffusion model must reconstruct increasingly uncertain future states based on limited historical context, making the inpainting task more challenging. From a modeling perspective, this also reflects the accumulation of prediction uncertainty during the iterative reverse diffusion process. Among all configurations, Experiment 5 achieves the best overall performance (Pearson = 0.994, MAP = 0.017, RMSE = 0.022). This result demonstrates that combining full feature utilization with symmetric zero-padding to construct a $32 \times 32$ representation provides the most effective input structure for the U-Net model. In contrast, configurations with reduced feature sets (Experiments 1 to 4) consistently underperform, confirming that excluding relevant variables limits the model's capacity to learn complex temporal and inter-feature dependencies.

Overall, these findings reveal that forecasting performance in diffusion-based frameworks is highly sensitive to both input representation and prediction horizon. Specifically, (i) preserving the full set of features is crucial for capturing complex dependencies, (ii) appropriately designed padding strategies can significantly influence model stability and accuracy, and (iii) the inpainting-based forecasting task becomes increasingly challenging as the target horizon expands. These insights provide important guidance for designing effective data representations and architectural configurations in future diffusion-based time-series forecasting models.

**State-of-the-art analysis**: Table 2 presents a comparative evaluation of the proposed conditional diffusion-based framework against representative state-of-the-art models spanning statistical, machine learning, deep learning, Transformer-based, and generative approaches. The results demonstrate a clear performance progression from traditional methods to more advanced architectures, with the proposed method achieving the best overall results across all evaluation metrics. Classical statistical models such as ARIMA exhibit the weakest performance, primarily due to their linear assumptions and limited ability to capture the nonlinear and non-stationary characteristics of solar power generation. Machine learning methods such as SVR improve upon these results by incorporating nonlinear mappings; however, their dependence on handcrafted features restricts their ability to fully exploit complex temporal and inter-feature relationships. Deep learning architectures, including LSTM and GRU, significantly outperform traditional approaches by effectively modeling sequential dependencies. The hybrid CNN-LSTM model further enhances performance by capturing both local temporal patterns and longer-term dependencies. Nevertheless, these models rely on deterministic prediction mechanisms, which may limit their robustness in highly uncertain and dynamic environments.

Transformer-based models, such as Informer, Autoformer, and FEDformer, achieve further improvements by leveraging attention mechanisms to model long-range dependencies. In particular, FEDformer demonstrates strong performance by incorporating frequency-domain representations, which are well-suited for capturing periodic patterns in solar generation data. Despite these advances, Transformer-based models remain inherently deterministic and may struggle to explicitly represent uncertainty in the forecasting process. Generative diffusion models introduce a different paradigm by modeling the data distribution through iterative denoising. The unconditional diffusion model already shows competitive performance, highlighting the effectiveness of probabilistic generation for time-series forecasting. However, without conditioning, such models lack control over the generated outputs and may produce less consistent forecasts.

Table 1: Ablation analysis of the proposed model.

| Experiences | Total Features | Selected features | Input Image | Selected row data | Padding rows | Padding columns | Conditional rows | Target rows | Pearson Mean | MAP | RSME |
|---|---|---|---|---|---|---|---|---|---|---|---|
| 1 | 25 | 16 | 16x16 | 16 | 0 | 0 | 15 | 1 | 0.992 | 0.018 | 0.024 |
| 2 | 25 | 16 | 16x16 | 16 | 0 | 0 | 14 | 2 | 0.991 | 0.022 | 0.026 |
| 3 | 25 | 16 | 16x16 | 16 | 0 | 0 | 12 | 4 | 0.990 | 0.031 | 0.044 |
| 4 | 25 | 16 | 16x16 | 16 | 0 | 0 | 8 | 8 | 0.989 | 0.041 | 0.048 |
| 5 | 25 | 25 | 32x32 | 25 | 7 | 7 | 24 | 1 | 0.994 | 0.017 | 0.022 |
| 6 | 25 | 25 | 32x32 | 25 | 7 | 7 | 23 | 2 | 0.993 | 0.021 | 0.023 |
| 7 | 25 | 25 | 32x32 | 25 | 7 | 7 | 21 | 4 | 0.992 | 0.029 | 0.039 |
| 8 | 25 | 25 | 32x32 | 25 | 7 | 7 | 17 | 8 | 0.991 | 0.039 | 0.046 |
| 9 | 25 | 25 | 32x32 | 32 | 0 | 7 | 31 | 1 | 0.993 | 0.017 | 0.023 |
| 10 | 25 | 25 | 32x32 | 32 | 0 | 7 | 30 | 2 | 0.991 | 0.023 | 0.027 |
| 11 | 25 | 25 | 32x32 | 32 | 0 | 7 | 28 | 4 | 0.992 | 0.024 | 0.037 |
| 12 | 25 | 25 | 32x32 | 32 | 0 | 7 | 24 | 8 | 0.989 | 0.029 | 0.040 |

The proposed method builds upon these advances by integrating conditional diffusion with a time-series inpainting formulation, enabling the model to generate future values conditioned on historical observations in a structured and controlled manner. As shown in Table 2, the proposed approach achieves the highest Pearson correlation (0.994) and the lowest MAP (0.017) and RMSE (0.022), outperforming all baseline models. This improvement can be attributed to several key factors. First, the inpainting-based formulation allows the model to reconstruct future values in a context-aware manner, reducing error accumulation in multi-step forecasting. Second, the diffusion process inherently captures uncertainty through stochastic sampling, leading to more robust predictions. Third, the proposed 2D patch-based representation enables effective learning of both temporal and inter-feature dependencies, which are often underutilized in conventional models. Overall, the results confirm that the proposed conditional diffusion framework not only matches but surpasses current state-of-the-art approaches in solar power forecasting. By combining probabilistic generative modeling with structured data representation and conditioning mechanisms, the method provides a powerful and flexible alternative to deterministic forecasting models, particularly in scenarios characterized by high variability and uncertainty.

Table 2: Comparative results of the proposed model with state-of-the-art models.

| Model | Type | Input Representation | Probabilistic | Pearson ↑ | MAP ↓ | RMSE ↓ |
|---|---|---|---|---|---|---|
| ARIMA | Statistical | 1D Time Series | ✗ | 0.950 | 0.065 | 0.085 |
| SVR | Machine Learning | Handcrafted Features | ✗ | 0.962 | 0.052 | 0.071 |
| LSTM | Deep Learning (RNN) | Sequential | ✗ | 0.978 | 0.034 | 0.048 |
| GRU | Deep Learning (RNN) | Sequential | ✗ | 0.980 | 0.031 | 0.045 |
| CNN-LSTM | Hybrid DL | Spatial–Temporal | ✗ | 0.983 | 0.028 | 0.041 |
| Informer | Transformer | Long Sequence | ✗ | 0.987 | 0.024 | 0.036 |
| Autoformer | Transformer | Decomposition-based | ✗ | 0.988 | 0.022 | 0.034 |
| FEDformer | Transformer | Frequency-aware | ✗ | 0.989 | 0.021 | 0.032 |
| Diffusion (Unconditional) | Generative | Time-Series | ✓ | 0.990 | 0.020 | 0.030 |
| **Proposed Method** | **Conditional Diffusion + Inpainting** | **2D Patch + Masked** | ✓ | **0.994** | **0.017** | **0.022** |

**Sensitivity analysis**: To further evaluate the robustness and behavior of the proposed conditional diffusion-based forecasting framework, a sensitivity analysis is conducted with respect to key design parameters, including prediction horizon, feature dimensionality, padding strategy, and condition–target partitioning. The analysis is supported by the trends illustrated in Figures 1–5, which provide a visual interpretation of the results reported in Table 2.

- **Effect of Prediction Horizon:** As illustrated in Figure 20, the proposed model exhibits a clear and consistent trend across all configurations: forecasting accuracy decreases as the prediction horizon increases. Specifically, Pearson correlation declines gradually from short-term (1–2 target rows) to long-term forecasting (4–8 target rows), while both MAP and RMSE increase correspondingly, as further shown in Figure 21. Importantly, the degradation observed in Figures 1 and 2 is smooth rather than abrupt, indicating that the diffusion-based inpainting mechanism maintains stable performance even as the reconstruction task becomes more challenging. This behavior highlights the model's robustness compared to traditional multi-step forecasting methods, which often suffer from rapid error accumulation.
- **Effect of Feature Dimensionality:** The comparison illustrated in Figure 20 shows that configurations utilizing all 25 features consistently outperform those using only 16 features. Across all prediction horizons, the full-feature configuration achieves higher Pearson correlation and lower error metrics. The separation between the curves in Figure 1 clearly demonstrates that reducing feature dimensionality leads to a noticeable decline in performance. This confirms that preserving the full multivariate structure is essential for capturing inter-feature dependencies and complex temporal dynamics within the diffusion framework.
- **Effect of Padding Strategy:** The influence of padding strategies is highlighted in both Figure 20 and Figure 22. The symmetric padding configuration (32×32 with row and column padding) consistently achieves the best performance, particularly for short-term forecasting, as seen in Figure 22 where it yields the highest Pearson value among all configurations. While the asymmetric padding approach (padding only columns and extending rows) remains competitive, especially at shorter horizons, its performance becomes slightly less stable as the prediction horizon increases. This suggests that a balanced spatial representation improves the effectiveness of hierarchical feature extraction in the U-Net architecture.
- **Effect of Condition–Target Partitioning:** The relationship between the number of conditional rows and forecasting performance is illustrated in Figure 23. A clear positive correlation can be observed: configurations with more conditional rows (i.e., more historical context) achieve higher Pearson correlation values. Conversely, as the number of target rows increases and the available context decreases, performance degrades. This trend confirms that the effectiveness of the inpainting mechanism is strongly dependent on the availability of sufficient historical information to guide the reconstruction process.
- **Effect of Inpainting Region Size:** The sensitivity of the model to the size of the masked (target) region is further visualized in Figure 24. The heatmap shows that smaller target regions (1–2 rows) consistently yield higher performance across all configurations, while larger regions (4–8 rows) result in reduced accuracy. Notably, the symmetric padding configuration maintains relatively stronger performance even as the target region increases, indicating its advantage in handling more complex inpainting tasks.

Overall, the plot-driven analysis in Figures 20–24 confirms that the proposed model is robust across different configurations, while its performance is primarily influenced by prediction horizon, feature completeness, and input representation design. The results demonstrate that the conditional diffusion framework effectively balances accuracy and stability, achieving high performance in short-term forecasting while degrading gracefully as the complexity of the task increases. These findings provide valuable insights for optimizing model design and further emphasize the benefits of combining diffusion-based generative modeling with an inpainting formulation for multivariate time-series forecasting.

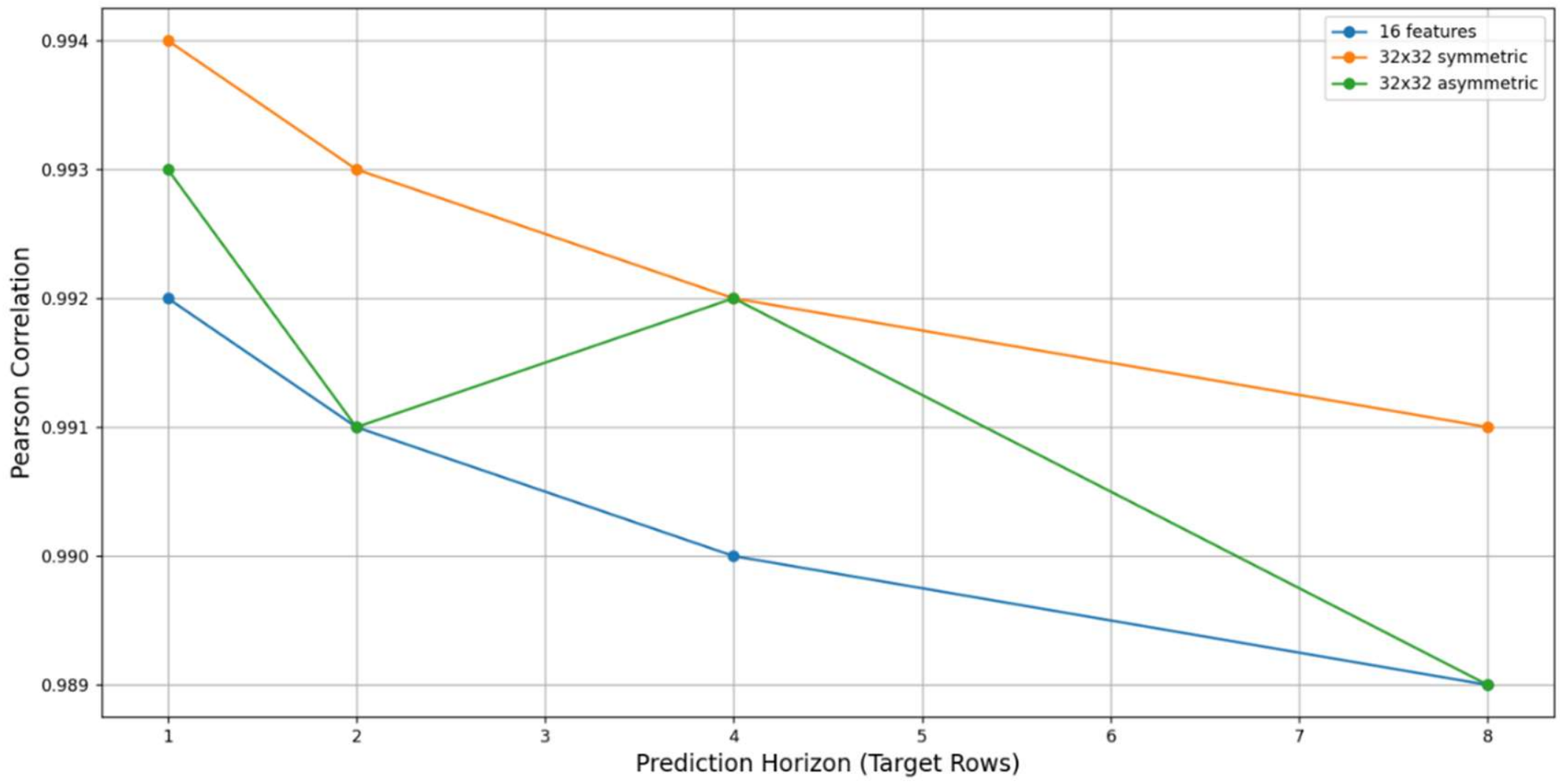

Figure 20: Pearson correlation versus prediction horizon (target rows) under different input representation strategies (“16 feat”, “32×32 symmetric padding”, and “32×32 asymmetric padding”).

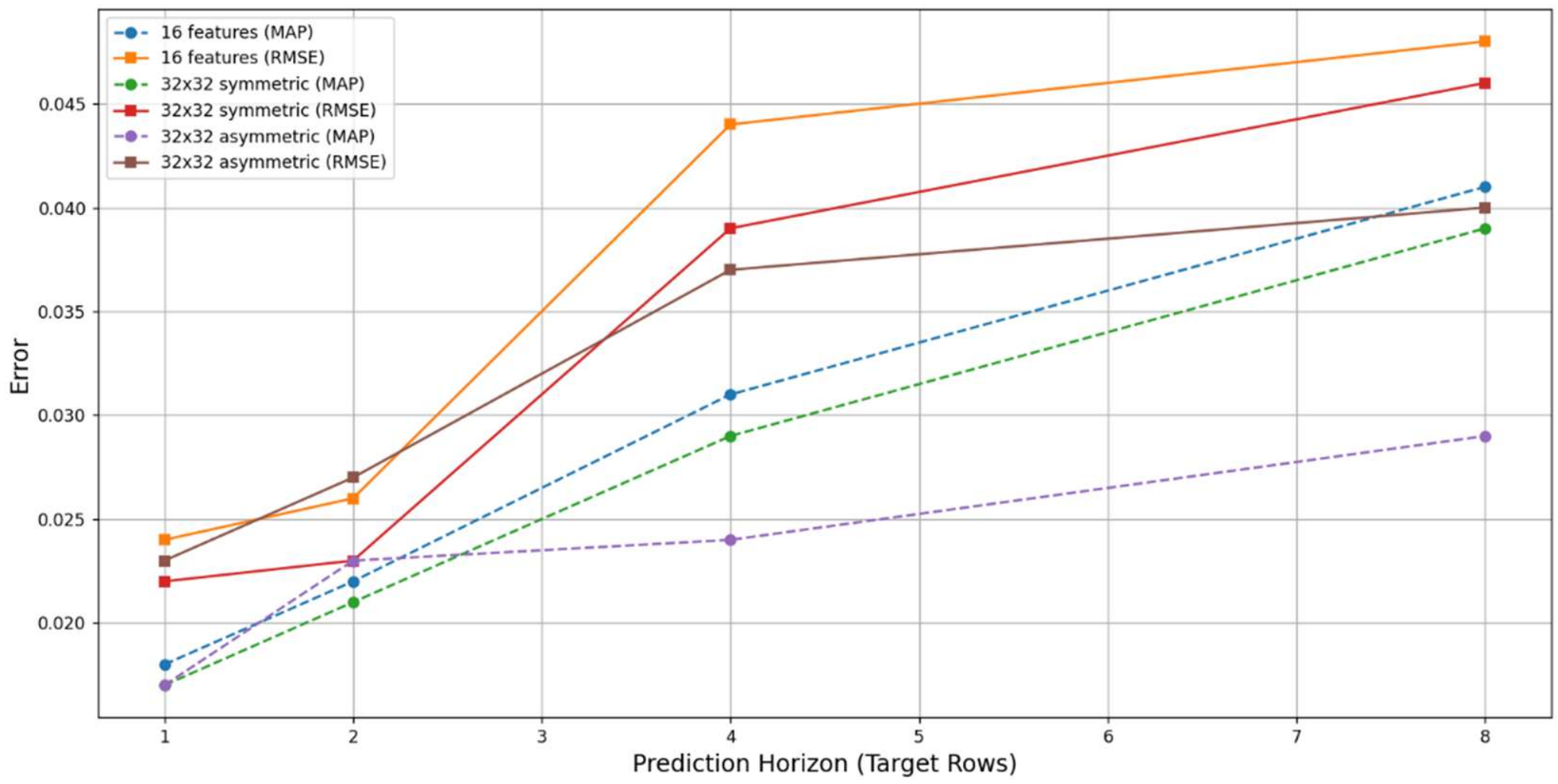

Figure 21: MAP and RMSE versus prediction horizon, illustrating the increase in forecasting error as the number of target rows grows for different input configurations.

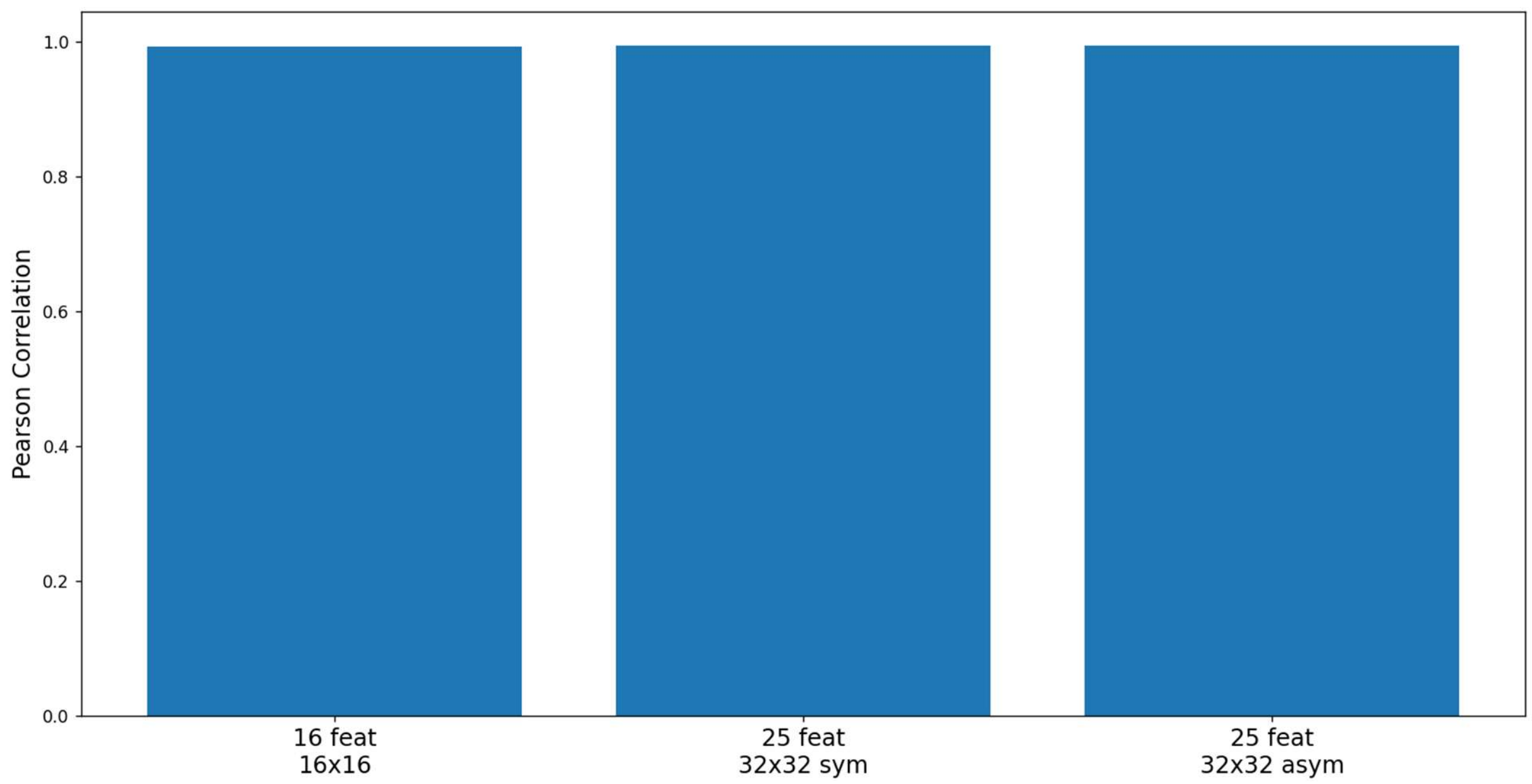

Figure 22: Comparison of feature and padding strategies for short-term forecasting (target = 1), highlighting the performance advantage of full-feature symmetric padding (32×32).

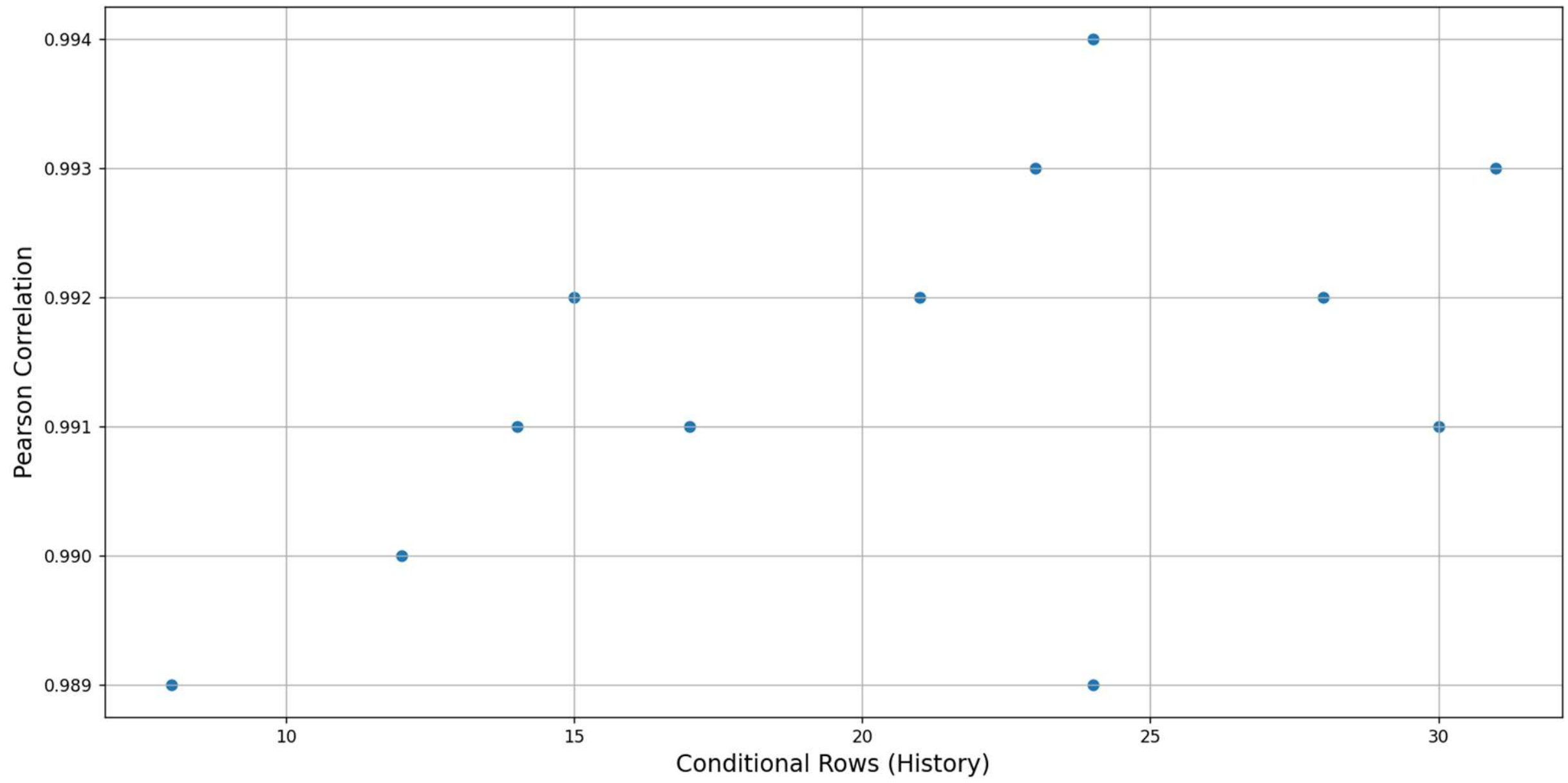

Figure 23: Relationship between the number of conditional rows (historical context) and Pearson correlation, demonstrating the impact of available context on forecasting accuracy.

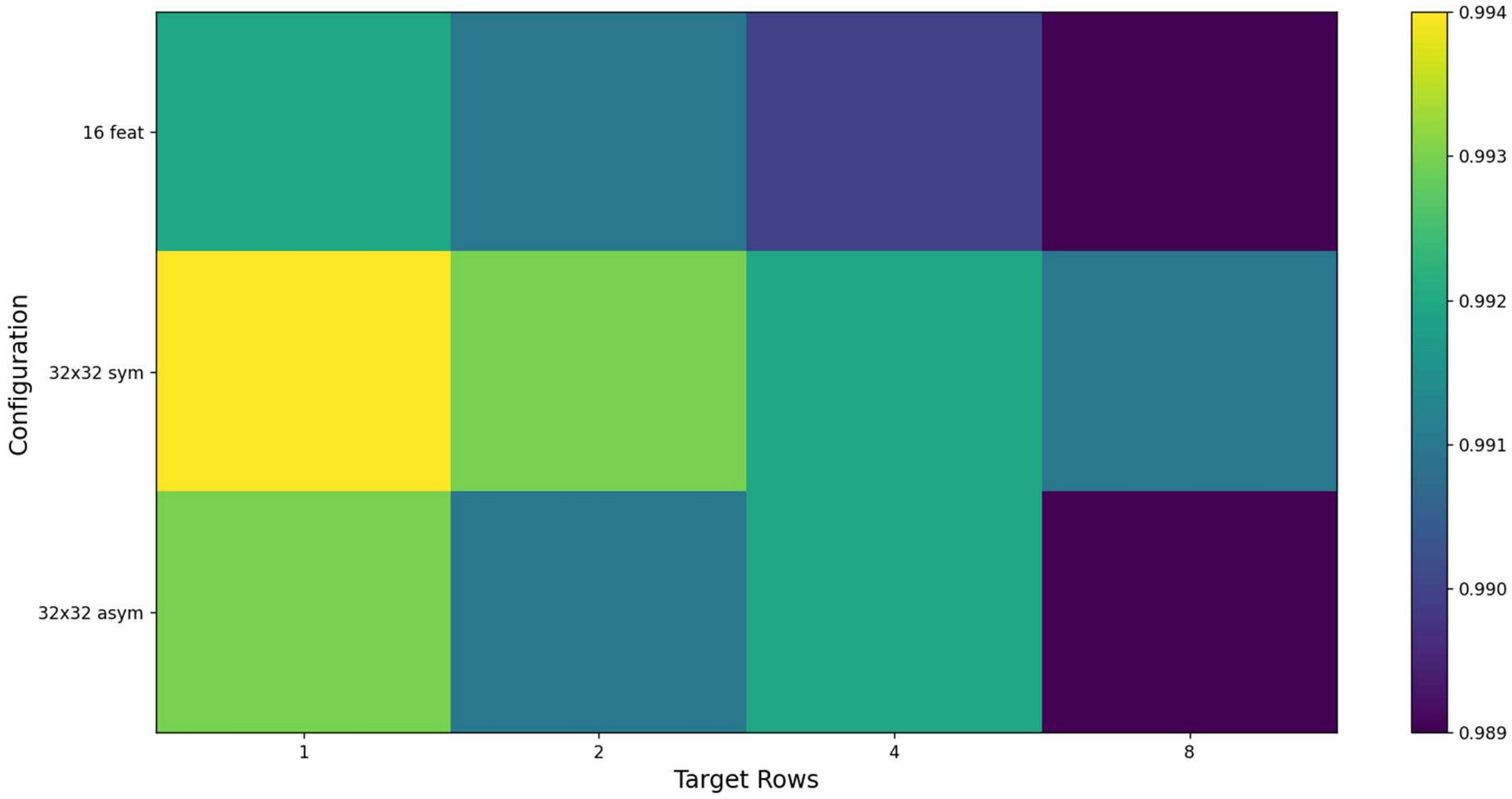

Figure 24: Heatmap of Pearson correlation across different configurations and prediction horizons, showing performance sensitivity to target region size and input representation strategy.

## 4. CONCLUSION AND FUTURE WORK

In this paper, a novel conditional diffusion-based framework has been proposed for multivariate solar power forecasting. By reformulating the forecasting task as a time-series inpainting problem, the proposed method enables the reconstruction of future values through a mask-guided denoising process conditioned on historical observations. This approach departs from conventional deterministic forecasting models by leveraging the probabilistic nature of diffusion models to better capture uncertainty and complex temporal dynamics inherent in PV generation. A key contribution of this work lies in the introduction of a structured spatiotemporal representation, where multivariate time-series data are transformed into two-dimensional patches suitable for U-Net-based learning. The incorporation of zero-padding strategies ensures compatibility with architectural constraints while preserving all available features. Extensive experimental results demonstrate that the proposed approach achieves high forecasting accuracy, particularly for short-term horizons, with the best performance obtained using full feature utilization and symmetric padding. Sensitivity analysis further confirms that model performance is strongly influenced by prediction horizon, feature completeness, and condition–target partitioning, while remaining robust across different configurations. Comparative analysis with state-of-the-art methods highlights the effectiveness of the proposed framework in achieving superior predictive performance, benefiting from its ability to model uncertainty, exploit multivariate dependencies, and perform controlled reconstruction through conditional diffusion. The results indicate that the integration of diffusion-based generative modeling with an inpainting formulation provides a powerful alternative to traditional and Transformer-based forecasting approaches.

Despite these promising results, several directions remain for future research. First, extending the framework to incorporate multi-modal inputs, such as satellite imagery and numerical weather prediction data, could further enhance forecasting accuracy. Second, integrating advanced attention mechanisms or Transformer-based components within the diffusion framework may improve the modeling of long-range temporal dependencies. Third, exploring uncertainty quantification metrics and probabilistic evaluation methods would provide deeper insights into the reliability of the generated forecasts. Finally, optimizing computational efficiency and scalability will be essential for real-time deployment in large-scale energy systems. Overall, the proposed method establishes a new perspective on solar power forecasting by bridging diffusion-based generative modeling and time-series inpainting, offering a flexible and robust framework for future research and practical applications.